\documentclass{sbc2023}%

\usepackage{graphicx}
\usepackage[misc,geometry]{ifsym} 
\usepackage{color}
\usepackage{hyperref} 
\usepackage{amsmath} %
\usepackage{amstext} %
\usepackage{aas_macros}
\usepackage[bottom]{footmisc}
\usepackage{supertabular}
\usepackage{afterpage}
\usepackage{url}
\usepackage{pifont}
\usepackage{multicol}
\usepackage{multirow}
\usepackage{bm}

\usepackage{subfig}

\newcommand{\miniscule}{\fontsize{4}{5}\selectfont}
\usepackage{lipsum}  
\usepackage{algorithmicx}
\usepackage[ruled]{algorithm}
\usepackage{algpseudocode}
\usepackage[table,dvipsnames]{xcolor}
\usepackage{booktabs}
\usepackage{array}
\usepackage[acronym]{glossaries}
\usepackage{graphicx}
\usepackage{xspace}
\usepackage{tikz}

\usepackage{tikz}
\usetikzlibrary{fit,positioning}
\usetikzlibrary{backgrounds}
\usetikzlibrary{positioning, arrows.meta, shapes.misc}

\setcitestyle{square}

\definecolor{orcidlogo}{rgb}{0.37,0.48,0.13}
\definecolor{unilogo}{rgb}{0.16, 0.26, 0.58}
\definecolor{maillogo}{rgb}{0.58, 0.16, 0.26}
\definecolor{darkblue}{rgb}{0.0,0.0,0.0}
\hypersetup{colorlinks,breaklinks,
            linkcolor=darkblue,urlcolor=darkblue,
            anchorcolor=darkblue,citecolor=darkblue}

\newacronymstyle{long-short-br}
{%
  \GlsUseAcrEntryDispStyle{long-short}%
}%
{%
  \GlsUseAcrStyleDefs{long-short}%
}
\setacronymstyle{long-short-br}

\newcommand\major[1]{#1} %
\newenvironment{majorblock}{\begingroup}{\endgroup}

\jtitle{This is the author-prepared version of a paper accepted for publication in the \textit{Journal of the Brazilian Computer Society}.\\
The version of record is available in the SBC OpenLib digital library under DOI: \href{https://doi.org/10.5753/jbcs.2026.5884}{\textcolor{blue}{10.5753/jbcs.2026.5884}}.}

\newcommand{\papertitle}{Robust Face Super-Resolution and Recognition Through Multi-Feature Aggregation in Diffusion Models}

\title[Robust Face Super-Resolution and Recognition Through Multi-Feature\\Aggregation in Diffusion Models]{\papertitle}

\author[dos Santos et al. 2026]{
\affil{\textbf{Marcelo dos Santos}\textsuperscript{*}~~[~\textbf{Federal University of Paraná}~|\href{mailto:msantos@inf.ufpr.br}{~\textbf{\textit{msantos@inf.ufpr.br}}}~]}

\affil{\textbf{Rayson Laroca}~~[~\textbf{Pontifical Catholic University of Paraná, Federal University of Paraná}~|\href{mailto:rayson@ppgia.pucpr.br}{~\textbf{\textit{rayson@ppgia.pucpr.br}}}~]}

\affil{\textbf{João Carlos Raposo Neves}~~[~\textbf{University of Beira Interior}~|\href{mailto:jcneves@di.ubi.pt}{~\textbf{\textit{jcneves@ubi.pt}}}~]}

\affil{\textbf{David Menotti}~~[~\textbf{Federal University of Paraná}~|\href{mailto:menotti@inf.ufpr.br}{~\textbf{\textit{menotti@inf.ufpr.br}}}~]}

}

\newcommand*{\RL}[2][]{\textcolor{Rhodamine}{[\textbf{\ifthenelse{\equal{#1}{}}{RL}{RL(#1)}}: #2]}}

\begin{document}

\begin{frontmatter}
\maketitle

\begin{mail}
Federal University of Paraná, Department of Informatics, Curitiba-PR, 81531-970, Brazil
\end{mail}

\newacronym{auc}{AUC}{Area Under the Curve}
\newacronym{cs}{CS}{Cosine Similarity}
\newacronym{ddpm}{DDPM}{Denoising Diffusion Probabilistic Models}
\newacronym{fc}{FC}{Feature Combiner}
\newacronym{fasr}{FASR}{Feature Aggregation Super-Resolution}
\newacronym{ffhq}{FFHQ}{Flickr-Faces-HQ}
\newacronym{gan}{GAN}{Generative Adversarial Network}
\newacronym{hr}{HR}{high-resolution}
\newacronym{mcmc}{MCMC}{Markov chain Monte Carlo}
\newacronym{mse}{MSE}{Mean Squared Error}
\newacronym{psnr}{PSNR}{peak signal-to-noise ratio}
\newacronym{smld}{SMLD}{Score matching with Langevin dynamics}
\newacronym{sr}{SR}{super-resolution}
\newacronym{lr}{LR}{low-resolution}
\newacronym{ssim}{SSIM}{structural similarity index measure}
\newacronym{sde}{SDE}{Stochastic Differential Equation}
\newacronym{sdes}{SDEs}{Stochastic Differential Equations}
\newacronym{sota}{SOTA}{state-of-the-art}
\newacronym{ve}{VE}{\major{Variance} Exploding}
\newacronym{vp}{VP}{\major{Variance} Preserving}
\newacronym{srname}{FASR++}{Feature Aggregation Super-Resolution}
\newcommand{\ffhq}{\gls*{ffhq}\xspace}
\newcommand{\gfpgan}{GFP-GAN\xspace}
\newcommand{\sparnet}{SPARNet\xspace}

\begin{abstract}
\textbf{Abstract.~}
\noindent 
Images acquired in surveillance environments often suffer from conditions such as low resolution, variations in pose, irregular illumination, and occlusions. Due to the low quality of these images, face recognition algorithms often struggle. This major limitation can be addressed by employing super-resolution techniques that enhance the details of the image. However, due to the high degree of difficulty of the problem, most super-resolution algorithms tend to cause distortions in the image and in the individual's identity. Thus, additional information must be incorporated into the processing to improve recognition robustness. In this regard, surveillance cameras can capture multiple images, even at low quality, and the data extracted from these images, such as consecutive video frames, can significantly enhance both super-resolution and facial recognition.
In this work, we introduce FASR++, a diffusion-model-based super-resolution algorithm. It leverages a reference low-resolution image and features extracted from multiple auxiliary low-quality images to generate a super-resolved output, minimizing distortions in the individual's identity.
Our approach recovers facial features without explicitly providing soft attributes or computing a function gradient to guide the reconstruction process. FASR++ generates high-quality images that can considerably improve performance in face recognition tasks when used as a pre-processing step.
We validate our approach on two standard face recognition datasets and attain state-of-the-art results for verification, face recognition, \major{and image quality metrics such as PSNR, SSIM, and~LPIPS.}
\end{abstract}


\end{frontmatter}

\section{Introduction}
\label{sec:introduction}

\glsresetall

 \begin{figure*}[!ht]
\centering
\includegraphics[width=0.98\linewidth]{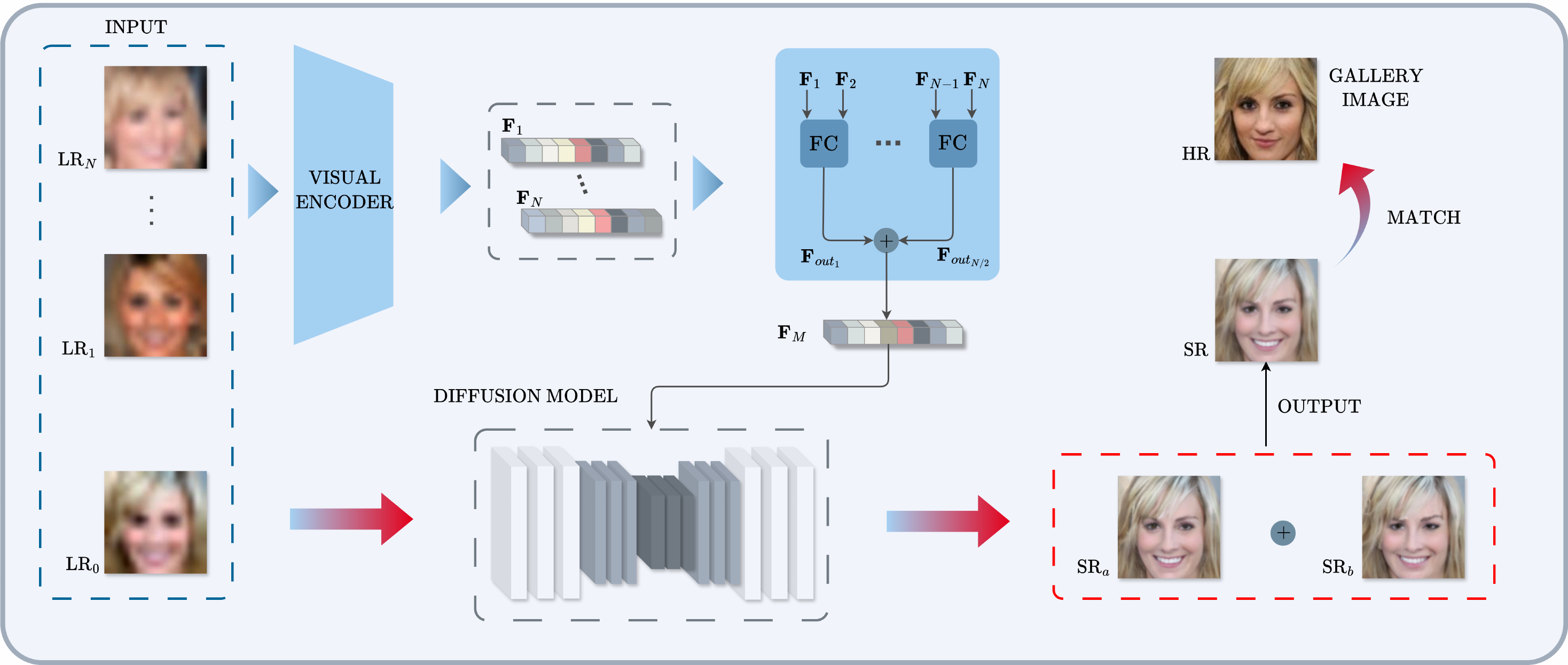}
\vspace{0.5mm}
\caption{\major{
\textbf{Overview of the proposed method.} 
At inference time, a set of $N{+}1$ low-resolution images is collected from an individual. 
The images $\text{LR}_1, \dots, \text{LR}_N$ are processed to extract feature representations $\mathbf{F}_1, \dots, \mathbf{F}_N$, 
which are then aggregated through an ensemble of Feature Combiner (FC) modules to produce the merged feature vector $\mathbf{F}_M$. 
The reference image $\text{LR}_0$ is jointly integrated with $\mathbf{F}_M$ into the diffusion model to generate two super-resolved outputs, SR$_a$ and SR$_b$. 
Their arithmetic mean yields the final super-resolved image SR, which is compared against a gallery of facial images for identity matching.
}}

\label{fig_overview}
\end{figure*}

In surveillance scenarios, the presence of noise, occlusions, variations in illumination, and varying poses poses a challenge even for \gls*{sota} face recognition algorithms, leading to a significant decline in performance  \citep{chag-surv-2016}. The most critical issue is the low-resolution and low-quality images acquired in real-world scenarios. Thus, the idea of utilizing \gls*{sr} algorithms as a pre-processing step for face recognition is not new \citep{emil2011} and arose as a natural solution to generate images with higher quality.
However, the super-resolution problem is inherently ill-posed, making the recovery of fine details and a reliable identity quite challenging \citep{ill-2002, jiang2021deep, nascimento2022combining, nascimento2024enhancing}.
In this sense, some works have focused on performing super-resolution by leveraging soft attributes such as eyeglasses, beards, mustaches, gender, and others as an additional source of information to reduce ambiguity and provide more robust results \citep{lee2018attribute, yu2018super,lu2018attribute,visapp24sr}. %
Nevertheless, facial attributes are often indistinct in \gls*{lr} images, making reliable identification challenging.
Also, obtaining these attributes requires a classifier or manual extraction, which is not very efficient~\citep{visapp24sr}.
However, many characteristics can be employed to assist super-resolution algorithms.
These include subtle facial proportions, skin textures, shapes, and other high-level, more abstract features that are not easily labeled or categorized.

Given the scarcity of robust super-resolution methods capable of preserving identity, this work aims to address this gap by focusing on identity preservation. We develop %
\textit{\gls*{srname}}, a robust \gls*{sr} algorithm that recovers crucial features for face recognition. It is more effective because, in addition to the \gls*{lr} image, it also takes as input a reliable vector of facial features derived from a set of \gls*{lr} images, such as a series of video frames or independent photos of an individual. This new vector has a higher signal-to-noise ratio than each individual vector. We incorporate it into the network, merging its information with the \gls{lr} image to generate an \gls{sr} version.
In this way, our algorithm effectively recovers facial information from an image, yielding higher-quality results with minimal distortion of identity.
Notably, \gls*{srname} employs a diffusion model based on a \gls*{sde} and does not require a classifier to guide the reverse diffusion process.

A preliminary version of this work, introducing the \gls*{fasr} method, was published at the 2024~Conference on Graphics, Patterns and Images~(SIBGRAPI)~\citep{santos2024multi}.
This paper builds upon that work with the following key improvements:
(i)~we develop a neural network specifically designed to enhance feature integration in low-resolution images, enabling more effective fusion and recovery of high-frequency details.
A hypothesis test validates its ability to merge complementary features;
(ii)~we incorporate additional evaluation metrics, expanding beyond recognition and verification performance to include image quality assessments such as 
\major{Structural Similarity Index Measure} (SSIM), \major{Peak Signal-to-Noise Ratio} (PSNR), and \major{Learned Perceptual Image Patch Similarity (LPIPS)~\citep{zhang2018unreasonable}};
(iii)~we conduct a comprehensive ablation study, demonstrating that the proposed method significantly outperforms the original FASR and other baseline~models.

Our method's effectiveness has been validated on the popular CelebA~\citep{liu2015faceattributes} and Quis-Campi~\citep{neves2018quis} datasets, and the key contributions are as follows:
\begin{itemize}
    \item We present a neural network that fuses and enhances low-resolution features, and we demonstrate its effectiveness in feature~integration; 
    \item We introduce FASR++, an improvement of FASR \citep{santos2024multi}, which utilizes the proposed neural network to generate high-quality merged features for assisting the reconstruction of super-resolution images in the diffusion~model;
    \item Our approach achieves superior qualitative results, producing more natural images with less distortion compared to \gls*{sota} super-resolution~algorithms;
    \item \major{Our quantitative results outperform \gls*{sota} algorithms in both image quality metrics, such as PSNR, SSIM and LPIPS, and identity-related metrics, including  \gls*{auc} in 1:1 verification, and accuracy in the 1:N~identification.}
\end{itemize}

\section{Related Work}
\label{sec:related}

\cite{sohl2015deep} introduced a generative model based on principles from non-equilibrium thermodynamics in their seminal work.
Two other influential studies in the field of diffusion models are Denoising Diffusion Probabilistic Models (DDPMs) \citep{ho2020denoising} and Score-Based Generative Models (SGMs) \citep{song2019generative, song2020improved}.
In \citep{song2021score}, DDPM and SGM are generalized for continuous time steps and noise levels using \gls*{sdes}, expanding the range of research possibilities in diffusion~models.

Due to the rapid evolution of diffusion models, various opportunities for their application have emerged. Recent works include the generation of audio, graphs, and shapes, as well as image synthesis, solutions of general inverse problems, and applications in medical images \citep{niu2020permutation, cai2020learning, ho2020denoising, song2019generative, song2021score, song2022solving}.
The full potential of diffusion models can also be leveraged through multi-domain data integration, such as text-to-image translation~\citep{saharia2022photorealistic} and image editing~\citep{zhang2023sine}.
Additionally, \cite{richter2023audio} combines audio-visual information for speech~enhancement.

\gls*{sr} is another important application of diffusion models and is investigated in this work.
In~\cite{saharia2023image}, an adaptation of the DDPM model produces high-quality \gls*{sr} images.
Similarly, SRDiff~\citep{li2022srdiff} employs diffusion models to estimate the difference between the original \gls*{lr} image and a \gls*{hr} image, resulting in an \gls*{sr} image.
In~\citep{dos2022face}, \glspl*{sde} were used to generate \gls*{sr} images.
Additionally, \cite{visapp24sr} performs \gls*{sr} by incorporating attribute information such as beard, gender, and the presence of eyeglasses to generate high-quality images.
However, their approach has the drawback that these attributes must be explicitly provided to the algorithm, which cannot be easily estimated in \gls*{lr}~images.

In \citep{suin2024diffuse}, an identity-preserving \gls*{sr} method was developed.
In both~\citep{visapp24sr} and \citep{suin2024diffuse}, a gradient must be calculated during the image reconstruction phase, which can increase computational cost. In this study, we develop an algorithm that restores image attributes by supplying a compact descriptor of facial features for the~algorithm.

Despite the impressive results achieved by diffusion models, their primary drawback is the high execution time caused by their iterative nature. Nevertheless, this issue is expected to be mitigated in the near future, as many studies focus on improving the computational efficiency of these methods. For a more in-depth discussion on accelerating sampling and enhancing efficiency in diffusion models, refer to~\citep{jolicoeur2021gotta, vahdat2021score, meng2023distillation}.

\section{Proposed Method}
\label{sec:Theoretical}

In this section, we present the general concept of the proposed method, followed by the description of a Feature Combiner module, the theoretical background on diffusion models formulated as SDEs, the model architecture, and the conditioning mechanisms based on low-resolution images, time, and feature embeddings.

\subsection{General Idea}

As previously noted, images captured in surveillance environments are often of low quality. Nevertheless, in certain instances, a video of a particular person can provide multiple low-resolution images that, when combined, can increase the valuable information necessary to recognize an individual.

In this work, we employ an \gls*{sr} algorithm to enhance a low-resolution image (LR$_0$), recovering useful information for face recognition (see Figure~\ref{fig_overview}).
We use a set of low-resolution auxiliary images LR$_1,\dots,$LR$_N$ of the same individual to extract a set of compact descriptors $\mathbf{F}_1,\dots,\mathbf{F}_{N}$. We train a {Feature Combiner} (FC) module (see Figure \ref{fig:fc}) to merge two vectors and recover the image's high-frequency information.
The vectors $\mathbf{F}_1,\dots,\mathbf{F}_{N}$ are combined through an ensemble of FCs to generate a representative compact descriptor  $\textbf{F}_M$. The reference low-resolution image LR$_0$ and the merged vector $\textbf{F}_M$ are input into a diffusion model, \major{which generates super-resolution images with minimal distortions in the identity}. Due to the stochasticity of diffusion models and the ill-posed nature of super-resolution problems, we can generate different solutions with small variations consistent with the same input LR$_0$. To increase the method's robustness, we generate two super-resolution images, SR$_a$ and SR$_b$, and combine them through an average to generate the final image, SR. This general framework leads to enhanced reliability concerning person identification.
Moreover, the algorithm successfully retrieves high-level features that might not be clearly visible but significantly enhance recognition accuracy and image quality.

\subsection{Feature Combiner (FC)}
\label{sec:fc}

\begin{figure}[!htb]
    \centering
    \resizebox{0.75\linewidth}{!}{\begin{tikzpicture}[
    font=\normalsize,
    node distance=8mm,
    box/.style={draw, very thick, rounded corners=3pt, minimum width=4.2cm, minimum height=1cm,
                text centered, fill=gray!20},
    outbox/.style={draw, very thick, rounded corners=3pt, minimum width=5.2cm, minimum height=1cm,
                  text centered, fill=gray!10},
    inout/.style={draw, very thick, rounded corners=3pt, minimum width=4.2cm, minimum height=1cm,
                 text centered, fill=cyan!8},
    arrow/.style={-{Latex[length=3mm,width=2mm]}, thick, draw=black!70}
]

\node[inout] (f1) {Input features $\mathbf{F}_i,\, \mathbf{F}_{i+1} \in \mathbb{R}^{512}$};

\node[outbox, below=of f1] (concat) {Concatenate $[\mathbf{F}_i,\, \mathbf{F}_{i+1}] \in \mathbb{R}^{1024}$};

\node[box, below=10mm of concat] (fc1) {Linear(1024→1024) + BN + ReLU + Dropout(0.2)};
\node[box, below=of fc1] (fc2) {Linear(1024→512) + BN + ReLU + Dropout(0.2)};
\node[box, below=of fc2] (fc3) {Linear(512→512)};

\node[outbox, below=16mm of fc3] (combine) {$\eta\cdot\delta([\mathbf{F}_{i},\mathbf{F}_{i+1}]) + \dfrac{\mathbf{F}_{i}+\mathbf{F}_{i+1}}{2}$};

\node[outbox, below=of combine] (norm) {$L_2$ normalization};

\node[inout, below=of norm] (out) {Output feature $\mathbf{F}_{\text{out},i'} \in \mathbb{R}^{512}$};

\begin{scope}[on background layer]
  \node[rectangle, draw, dashed, rounded corners,
        fill=teal!15, fill opacity=0.25, draw opacity=0.6,
        fit=(fc1)(fc2)(fc3),
        inner sep=14pt,
        label={[anchor=north west, xshift=0pt, yshift=0pt]below left:{}}]
        (groupbox) {};
\end{scope}

\draw[arrow] (f1.south) -- (concat.north);
\draw[arrow] (concat.south) -- (fc1.north);
\draw[arrow] (fc1.south) -- (fc2.north);
\draw[arrow] (fc2.south) -- (fc3.north);
\draw[arrow] (fc3.south) -- (combine.north)node[pos=0.62, right] {$\delta([\mathbf{F}_i,\mathbf{F}_{i+1}])$};
\draw[arrow] (combine.south) -- (norm.north);
\draw[arrow] (norm.south) -- (out.north);
\draw[arrow, rounded corners=8pt]   (f1.east)   -| ++(2.6,-1.0)   |-(combine.east); %

\end{tikzpicture}}
    \vspace{2.0mm}
\caption{\textbf{Architecture of the Feature Combiner.}
\major{The input features \( \mathbf{F}_i \) and \( \mathbf{F}_{i+1} \) are concatenated and processed by the neural network~$\delta$, 
which is composed of two fully connected layers with Batch Normalization (BN), ReLU activation, and Dropout (0.2), 
followed by a final linear layer. 
The resulting output of $\delta$ is combined with the mean of the input features to generate the final merged representation \( \mathbf{F}_{\text{out},i'} \).}}

    \label{fig:fc}
\end{figure}
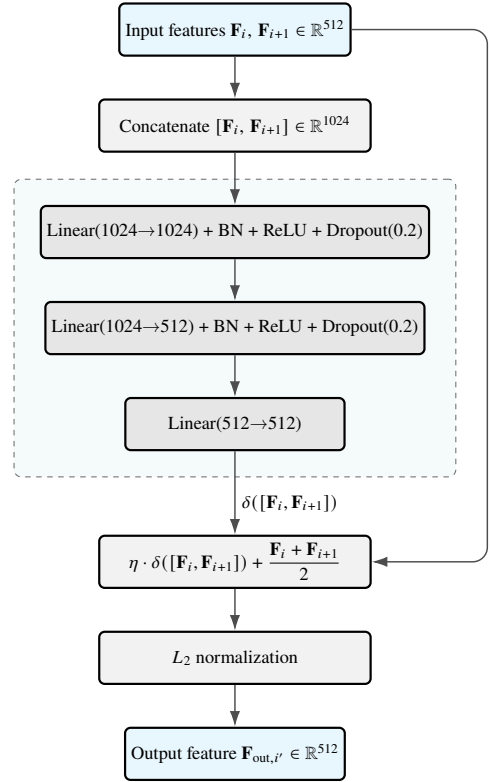

To train the \gls*{fc} module, we need a collection of low-resolution images and their high-resolution versions. Specifically, for a given identity, let $ \text{LR}_1, \dots, \text{LR}_N$ denote the set of $ N$ low-resolution images, and let $ \text{HR}_1, \dots, \text{HR}_N$ denote the respective set of high-resolution images. We then extract a set of compact face descriptors from these images using a pre-trained visual encoder to obtain the descriptors $\mathbf{F}_1,\dots,\mathbf{F}_{N},\mathbf{F}_1^\text{HR},\dots,\mathbf{F}_{N}^\text{HR}$ respectively.
The feature vectors from low-resolution images are combined in pairs using an ensemble of FC modules and posteriorly averaged to obtain an approximation of a reliable descriptor of a high-resolution image. 

\begin{majorblock}
More specifically, given two feature vectors \( \mathbf{F}_i \) and \( \mathbf{F}_{i+1} \), 
the FC module takes them as input and produces a single merged representation \( \mathbf{F}_{\text{out},i'} \), 
where \( i' = (i+1)/2 \), as illustrated in Figure~\ref{fig:fc}. 
The FC module consists of the mean of the input features and a refinement network~$\delta$. 
The mean operation was adopted as a baseline because the input low-resolution images, and consequently their extracted features, often contain noise and small distortions.  
Averaging allows the model to amplify the facial components that are common across different images of the same identity while attenuating uncorrelated noise and spurious variations.  
The FC then combines this mean with a learnable correction term generated by the neural network~$\delta$, which captures nonlinear relationships between the two feature vectors, modeling complex dependencies that cannot be represented by simple averaging.
Formally, the operation is
\begin{equation}
    \mathbf{F}_{\text{out},i'} = \text{FC}\left(\mathbf{F}_i, \mathbf{F}_{i+1}\right)
    = \frac{\mathbf{F}_i + \mathbf{F}_{i+1}}{2} + \eta \cdot \delta\left(\mathbf{F}_i, \mathbf{F}_{i+1}\right),
    \label{eqmerge}
\end{equation}
where \( \delta(\cdot) \) denotes the neural network and \(\eta \in \{0,1\}\) is a control parameter used to assess the influence of~\(\delta\) on the final results.

The architecture of~$\delta$, presented in Figure~\ref{fig:fc}, consists of three fully connected layers.  
The input is formed by concatenating the feature vectors $\mathbf{F}_i$ and $\mathbf{F}_{i+1}$, producing a 1024-dimensional representation.  
This concatenated vector is processed by two sequential linear transformations, each followed by Batch Normalization (BN) and ReLU activation, with a dropout rate of~0.2 applied after each nonlinear operation to improve generalization and reduce overfitting. The chosen dropout rate of~0.2 was determined through validation experiments and aligns with the typical values reported by \cite{labach2019survey}.
A final linear transformation produces a 512-dimensional feature refinement, which is combined with the mean $(\mathbf{F}_i + \mathbf{F}_{i+1}) / 2$ to generate the fused representation $\mathbf{F}_{\text{out},i'}$.
Finally, an $L_2$ normalization is applied to ensure unit-length embeddings.

\end{majorblock}

To optimize the network parameters, we employ a \textit{triplet loss}, aiming to maximize the similarity between positive matches while minimizing it for negative ones. In this framework, the anchor, positive and negative samples are:

\begin{itemize}
    \item \textbf{Anchor}: \( \mathbf{F}_{\text{out},i'}=\text{FC}\left(\mathbf{F}_i, \mathbf{F}_{i+1}\right) \), the combined feature representation generated by the FC module;
    \item \textbf{Positive sample}: The arithmetic mean of feature vectors from the respective HR image,~i.e.:
    \begin{equation}
        \frac{\mathbf{F}_i^\text{HR} + \mathbf{F}_{i+1}^\text{HR}}{2};
    \end{equation}
    \item \textbf{Negative sample}: A feature vector from a HR image of a different identity.
\end{itemize}

After the network~\(\delta\) is trained and each FC module effectively merges two feature vectors,  
an ensemble of FCs is employed to obtain the overall merged representation for a given individual.  
\begin{majorblock}
The final feature vector is computed by averaging all \(N/2\) merged outputs \(\mathbf{F}_{\text{out},i'}\),  
resulting in the final merged feature \(\mathbf{F}_M\), as expressed by
\begin{equation}
      \mathbf{F}_{M} = \frac{2}{N} \sum_{i'=1}^{N/2} \mathbf{F}_{\text{out},i'} 
      = \frac{1}{N} \sum_{i=1}^{N} \mathbf{F}_{i} 
      + \frac{2\eta}{N} \sum_{i=1}^{N/2} 
      \delta(\mathbf{F}_{2i-1}, \mathbf{F}_{2i}),
      \label{eqfm}
\end{equation}
for even \(N\).  
When \(N\) is odd, the last feature vector \(\mathbf{F}_{N}\) is averaged together with the \((N-1)/2\) merged features \(\mathbf{F}_{\text{out},i'}\).  
This formulation is general and applies to any number of input images. 
The averaging operation in Equation~\ref{eqfm} can be interpreted as analogous to the average pooling mechanism commonly employed in convolutional neural networks, where feature-wise averaging effectively reduces noise while preserving the dominant structural information.

In Equation~\ref{eqfm}, other operations, such as the maximum or the sum, could also be used. 
However, based on experimental analysis, we found that the mean yields superior results compared to these alternatives. 
Using the maximum may distort the features by emphasizing individual noisy components rather than the common underlying structure. 
A simple sum, in turn, would unnecessarily amplify the magnitude of the features as $N$ increases. 
In contrast, the mean provides a stable aggregation that yields merged features $\mathbf{F}_{\text{out},i'}$ closer to the target high-resolution ground truths, ensuring a coherent estimate of the underlying representation.
Since the individual feature vectors are normalized, the mean effectively balances their contributions without being dominated by any single component.
Moreover, residual noise components may still be present, and averaging serves as a natural denoising mechanism that increases the signal-to-noise ratio and stabilizes the overall representation across different image conditions.

From Equation~\ref{eqfm}, it follows that the mean of $\mathbf{F}_{\text{out},i'}$ is equivalent to a single global mean 
over all input features $\mathbf{F}_{i}$, plus an additional term involving $\delta(\cdot,\cdot)$ that represents the nonlinear refinement 
computed from sequential feature pairs by the Feature Combiner. 
Therefore, if another nonlinear function were more suitable for feature aggregation than the mean, the $\delta$ network would implicitly compensate for it.
\end{majorblock}

This formulation ensures that the learned representation effectively consolidates the features while preserving identity and enhancing discriminability among individuals.

\subsection{Theoretical Background}

In the context of image generation, diffusion models have two phases: forward diffusion and reverse diffusion.
During forward diffusion, Gaussian noise is added to the image, and a network is trained to predict this noise.
In reverse diffusion, an image composed purely of noise is iteratively denoised and transformed into an image that follows a distribution similar to the images in the training set.
If the diffusion procedure is continuous, it can be modeled using an~\gls*{sde}.

According to \cite{song2021score, anderson1982reverse}, a forward diffusion process $\{\mathbf{x}(t)\}_{t=0}^T$ and its reverse are, respectively, modeled using the following \glspl*{sde}:
\begin{equation}
    \mathrm{d}\mathbf{x}=\mathbf{f}(\mathbf{x},t)\mathrm{d}t+g(t)\mathrm{d}\mathbf{w},
    \label{eqsde}
\end{equation}
\begin{equation}
    \mathrm{d}\mathbf{x}=[\mathbf{f}(\mathbf{x},t)-g(t)^2\nabla_{\mathbf{x}}\log p_t(\mathbf{x})]\mathrm{d}t+g(t)\mathrm{d}\bar{\mathbf{w}},
    \label{revsde}
\end{equation}
where $\mathbf{f}(\mathbf{x},t)$ is the drift coefficient, $g(t)$ is a diffusion coefficient,  $\mathbf{w}$ and $\bar{\mathbf{w}}$ are Wiener process (\major{the latter runs backward in time}) and $p_t$ is the probability density of $\mathbf{x}(t)$.  \cite{sdeplaten, sarkka2019applied} supply more details about Itô \gls*{sde}s and the Wiener~process.

Here, we consider $\mathbf{x}_t$ as an image to be denoised. At $t=0$, the noise level in the image is zero, and at $t=T$, the noise is at its maximum, and there is no information on the image. 
\begin{majorblock}
To obtain a super-resolved image, we solve the reverse diffusion process defined in Equation~\ref{revsde}. 
For this purpose, a deep neural network $s_\theta$ is employed to approximate the score function $\nabla_{\mathbf{x}}\log p_t(\mathbf{x})$. 
During this reverse process, $s_\theta$ is conditioned on both the reference low-resolution image $\text{LR}_0$, denoted by $\mathbf{y}$, and the merged feature vector $\mathbf{F}_M$, which provides complementary guidance.
\end{majorblock}

The training of the neural network  $s_{\theta}$ %
is achieved by  optimizing the following loss function~\citep{vincent2011connection}: 
\begin{majorblock}
\begin{align}
     \min_{\theta}
    \mathbb{E}_{t\sim \mathcal{U}[0,T]}
   \mathbb{E}_{\mathbf{x}(0)\sim p(\mathbf{x}(0))}\mathbb{E}_{\mathbf{x}(t)\sim p_t(\mathbf{x}(t)|\mathbf{x}(0))}
    \big[\lambda(t)\quad\nonumber\\ \times\, \|s_\theta(\mathbf{x}(t),\mathbf{y},\mathbf{F}_\mathbf{y}^\text{HR},t)-\nabla_{\mathbf{x}(t)}\log p_{}(\mathbf{x}(t)|\mathbf{x}(0))\|_2^2\big],
    \label{eqloss}
\end{align}
where $\lambda(t)$ is a positive weighting function, $p_{}(\mathbf{x}(t)|\mathbf{x}(0))$ is the transition kernel from $\mathbf{x}(0)$ to $\mathbf{x}(t)$ and $\mathbf{F}_{\mathbf{y}}^{\text{HR}}$ represents the ground-truth features extracted from the high-resolution version of $\mathbf{y}$.  %
\end{majorblock}

Here, we use the  \gls*{ve}  case described in~\citep{song2021score} with $\mathbf{f}(\mathbf{x},t)$ and $g(t)$ given respectively by:
\begin{equation}
\mathbf{f}(\mathbf{x},t)=\mathbf{0},\quad g(t)=\sqrt{\frac{\mathrm{d}\sigma^2(t)}{\mathrm{d}t}} \, ,    
\end{equation}
\noindent where $\sigma(t)=\sigma_{min}\left(\sigma_{max}/\sigma_{min}\right)^t$ denotes the noise level of the image at the time $t$.

For $\mathbf{f}(\mathbf{x},t)$ and $g(t)$ described above, the mean and variance of $p(\mathbf{x}(t)|\mathbf{x}(0))$ are given by~\citep{song2021score}:
\begin{equation}
\pmb{\mu}(t) = \mathbf{x}(0), \,\, \pmb{\Sigma}(t)=[\sigma^2(t)-\sigma^2(0)]\mathbf{I}.  
\label{eq_mean}
\end{equation}
Thus, we can analytically compute $\nabla_{\mathbf{x}}\log p(\mathbf{x}(t)|\mathbf{x}(0))$ in Equation~\ref{eqloss},  allowing for efficient model training.
Once the network well estimates the gradient, we generate an SR image~$\mathbf{x}(0)$ by changing  $\nabla_{\mathbf{x}}\log p_t(\mathbf{x})$ by \major{$s_\theta(\mathbf{x}(t),\mathbf{y},\mathbf{F}_M,t)$} in the reverse process (Equation~\ref{revsde}) and solving it from $t=T$ to $t=0$ using the Euler-Maruyama method~\citep{sdeplaten, sarkka2019applied}.

\begin{majorblock}
\subsection{Diffusion Model Architecture}
\label{sec:architecture}
The proposed model adopts the NCSN++~\citep{song2021score} backbone, implemented as a U-Net~\citep{ronneberger2015u} with residual and attention blocks distributed in multiple resolutions.
The network comprises seven resolution levels ($128\!\to\!64\!\to\!32\!\to\!16\!\to\!8\!\to\!4\!\to\!2$),
with channel widths scaled by $(1, 1, 2, 2, 2, 2, 2)$ relative to an initial number of 128 feature maps.
Each level contains two residual blocks, followed by either downsampling or upsampling operations between successive resolutions.

Each residual block follows a standard design consisting of Group Normalization, a SiLU nonlinearity, and two $3\times3$ convolutional layers.  
A dropout rate of $0.1$ is applied within each block to improve regularization, while a $1\times1$ convolution is used for channel projection whenever the input and output dimensions differ.  
In addition, self-attention layers are inserted at the $16\times16$ resolution to capture long-range spatial dependencies across feature maps~\citep{vaswani2017attention}.

The network is conditioned on three signals: the low-resolution input image, the time embedding, and the merged feature vector. In
 each residual block, temporal and feature signals are injected through independent dense layers after SiLU activation, as described in Subsection~\ref{sec:conditioning}.

Finally, a $3\times3$ convolution maps the feature maps back to the RGB space.  
This configuration, combining residual, attention and multi-scale conditioning mechanisms, improves representational capacity and training stability~\citep{he2016deep,wang2017residual,liang2021swinir}.
\end{majorblock}

\subsection{Model Conditioning}
\label{sec:conditioning}
\begin{figure}[!htb]
\centering
\resizebox{0.99\linewidth}{!}{
\includegraphics[width=0.99\linewidth]{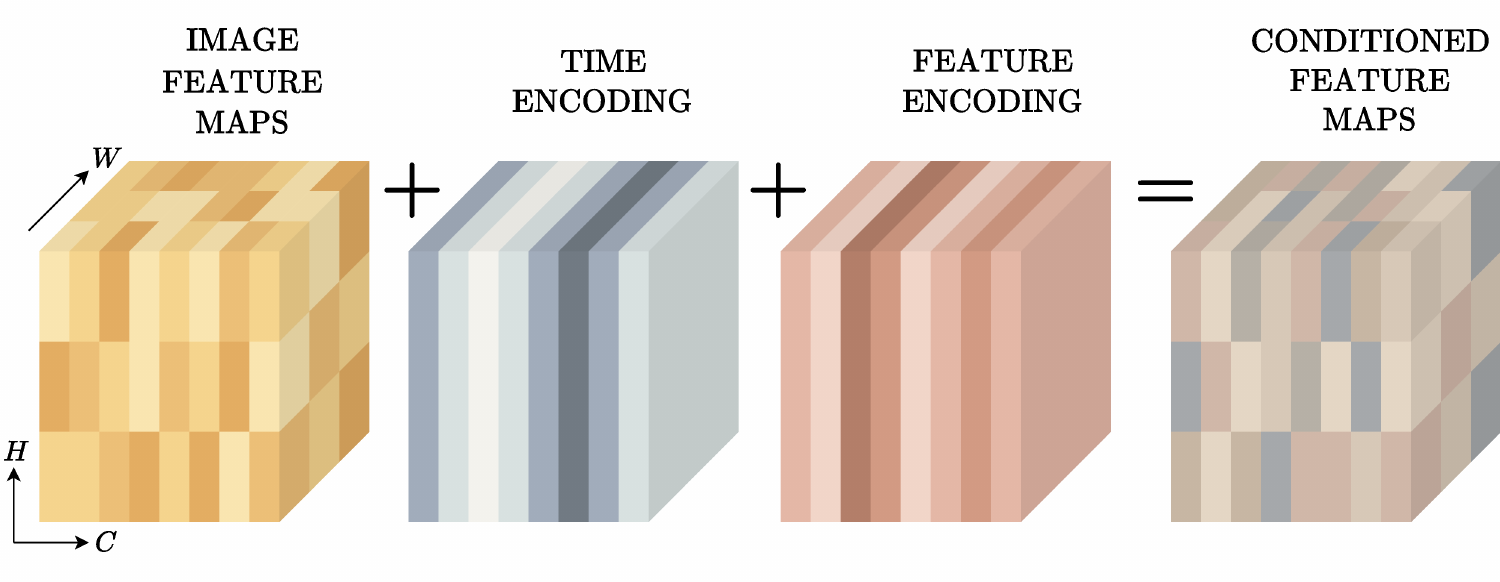}
}

\vspace{0.1mm}
\caption{\major{\textbf{Time and features encoding.} 
   Illustration of the conditioning mechanism in the proposed diffusion model.
The first block represents the {image feature maps} of shape ${(B, C, H, W)}$, where $B$ is the batch size. In this example, we illustrate a single element from the batch for clarity.
The time encoding and feature encoding tensors are broadcast along the spatial dimensions so that their values remain constant across $H$ and $W$, enabling element-wise addition with the {image feature maps}.
The final block represents the {feature maps conditioned on time and feature embeddings}, which are subsequently processed by the following layers of the network.
}}
\label{fig_embed}
\end{figure}

\begin{majorblock}
To guide the diffusion process, our model employs three complementary conditioning mechanisms:  
$(i)$~the low-resolution image, which provides spatial and textural priors;  
$(ii)$~the time embedding, which encodes the current denoising stage; and  
$(iii)$~the merged feature vector, which conveys high-level identity information obtained from the ensemble of feature combiners.  
Each of these conditioning signals plays a distinct role in steering the generation process toward faithful and identity-preserving reconstructions.

\paragraph{LR Image Conditioning.}
To condition the model on low-resolution input, we follow a strategy similar to that adopted in~\citep{dos2022face,saharia2023image}, where the conditioning image is directly concatenated with the noisy input image along the channel dimension.  
Specifically, the low-resolution image $\mathbf{y}$ and the noisy image $\mathbf{x}_T$ (the sample being progressively denoised) are concatenated to form a six-channel tensor
\[
\text{concatenate}[\mathbf{y}, \mathbf{x}_T] \in \mathbb{R}^{6\times H\times W},
\]
where $H$ and $W$ denote the spatial dimensions of the image.  
This tensor serves as the input to the U-Net backbone, allowing the network to take advantage of spatial and textural cues from the LR image throughout the denoising process.

\paragraph{Time Conditioning.}
The temporal conditioning follows the strategy adopted in other diffusion models~\citep{song2021score}, where the diffusion timestep $t \in [0, 1]$ is mapped into a high-dimensional embedding using Gaussian Fourier features~\citep{tancik2020fourier}.  
For a given timestep ${t}$, the Fourier mapping is computed as
\begin{equation}
\mathbf{E}_t = \text{concatenate}[\sin(2\pi \bm{\omega} t), \; \cos(2\pi\bm{\omega} t)],
\end{equation}
where $\bm{\omega} \in \mathbb{R}^{d}$ is a fixed, non-trainable vector of random frequencies drawn from a normal distribution.  
This produces a time embedding $\mathbf{E}_t \in \mathbb{R}^{2d}$ (with $d = 256$ in our implementation, yielding 512 dimensions).  

The embedding $\mathbf{E}_t$ is then processed by a SiLU activation followed by a single linear layer that projects it into the same channel dimension $C$ as the U-Net feature maps.
To enable element-wise addition with the convolutional features, $\mathbf{E}_t$ is reshaped to $\mathbb{R}^{C\times1\times1}$ and broadcast across $H$ and $W$, ensuring that the same temporal modulation is applied uniformly to all spatial positions within each residual block.

Time conditioning plays a crucial role in diffusion models, as it allows the network to adapt its denoising behavior according to the current noise level along the diffusion trajectory, improving both temporal coherence and reconstruction~quality.

\paragraph{Feature Conditioning.}
Analogously, the model is also conditioned on the merged feature vector $\mathbf{F}_M$ obtained from the ensemble of feature combiners.  
This vector is processed through a SiLU activation followed by a single linear layer, projecting it into the same channel dimension $C$ as the network feature maps. Then it is reshaped to $\mathbb{R}^{C\times1\times1}$ and broadcast across $H$ and $W$ to allow element-wise addition with the U-Net feature maps.  
In this way, the conditioning provided by $\mathbf{F}_M$ acts as a global control signal, influencing all spatial locations uniformly while preserving the channel-wise semantics learned by the network.

\paragraph{Integration within Residual Blocks.}
We now describe how these conditioning signals are incorporated into the network.  
At each residual block of the U-Net, let \( h \) denote the intermediate feature map after normalization and activation.  
The temporal embedding and the merged feature vector are injected into these residual blocks through independent linear layers D$_t$ and D$_f$, as follows:
\begin{equation}
h \leftarrow h
+ \text{D}_t(\text{SiLU}(\mathbf{E}_t))_{\text{reshaped}}
+ \text{D}_f(\text{SiLU}(\mathbf{F}_M))_{\text{reshaped}}.
\end{equation}
The layers \( \text{D}_t \) and \( \text{D}_f \) project these embeddings into the same channel dimension as \(h\), and the resulting tensors are reshaped to match the spatial dimensions of the feature maps.  
This mechanism allows both the diffusion timestep and the identity-related information to condition the U-Net feature maps across multiple scales, as illustrated in Figure~\ref{fig_embed}.
\end{majorblock}
\paragraph{Qualitative Demonstration of Feature Conditioning.}

\begin{figure}[!htb]
\centering
\setlength{\tabcolsep}{1.2pt}
\resizebox{0.9\linewidth}{!}{ %
\begin{tabular}{cccc}
\includegraphics[width=15mm]{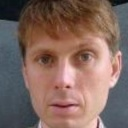} & 
\includegraphics[width=15mm]{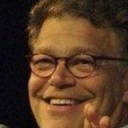} &\includegraphics[width=15mm]{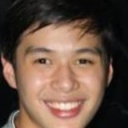}
&\includegraphics[width=15mm]{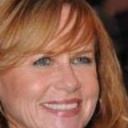}\\[-0.32ex]
\includegraphics[width=15mm]{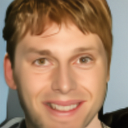} & 
\includegraphics[width=15mm]{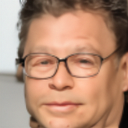}&
\includegraphics[width=15mm]{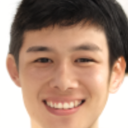}&
\includegraphics[width=15mm]{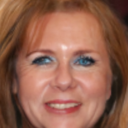}\\

\end{tabular}
}
\caption{\textbf{Efficacy of FASR++ for face reconstruction.} Extracted from \citep{santos2024multi}.
First row: original HR images from the CelebA dataset~\citep{liu2015faceattributes}.
Second row: synthetic HR images generated solely from the feature vectors extracted from the corresponding images in the first~row.}

\label{fig_from_fv}
\end{figure}

\major{To demonstrate the effectiveness of using the feature vector to generate an SR image,  
we trained our model with $\mathbf{y}=0$ in Equation~\ref{eqloss}, i.e., without using the LR image, and employed the ground-truth feature vectors to produce images, as illustrated in Figure~\ref{fig_from_fv}.
These images demonstrate the algorithm’s ability to reconstruct high-level features that encode identity-specific semantics, including facial geometry, landmark configuration, and characteristic texture patterns.
The low-resolution image, in turn, provides the global spatial layout and low-frequency information required to preserve geometric coherence, pose, and illumination consistency during reconstruction. It defines the structural basis for the generative process.  
When both signals are combined, they enable the model to generate visually convincing and identity-consistent faces.
}

\section{Experiments and Results}
\label{sec:Exp_and_Res}
This section describes the experimental setup and the results obtained on two different datasets.
Lastly, we present the ablation study and examine some extreme cases where the algorithm may fail.
\subsection{Experiments}
\label{sec_exp}

In this study, we explored four datasets: FFHQ~\citep{karras2019style}, \major{CASIA-WebFace}~\citep{yi2014learning}, CelebA~\citep{liu2015faceattributes}, and Quis-Campi~\citep{neves2018quis}, the latter of which originates from surveillance scenarios.
The FFHQ dataset was used for the diffusion model training, with $10^6$ training steps performed, \major{while the CASIA-WebFace dataset was specifically used to train the $\delta$ network. Further details on the $\delta$ training procedure are provided in the next section.}
CelebA was used to test our approach, with $500$ identities selected.
Each identity comprises multiple images, with one randomly chosen as the gallery image.
A second image is downsampled to create an \gls*{lr} probe image.
The remaining images were also downsampled and used to extract features, assisting the reconstruction of the LR probe image.

A complementary test to further validate our algorithm was conducted on a real-world scenario from the Quis-Campi dataset, where the images pose additional challenges for SR and face recognition algorithms~\citep{neves2018quis}.
We selected $90$ identities and used five downsampled images as probe images for each identity.
These images were then used to calculate an average feature vector, which was utilized to support the generation of the SR image.
In addition, the dataset already contains gallery images obtained in a controlled environment for each~identity.

The parameters controlling the noise level over time were set at $\sigma_{min}=0.001$ and $\sigma_{max} = 348$.
We worked with images of $128 \times 128$ pixels.
For producing LR images, we applied $8\times8$ downsampling followed by upsampling using bicubic interpolation to achieve a final size of $128 \times 128$ pixels.
We used $2{,}000$ steps to solve the SDE for image~reconstruction.

The feature vector used for both training the SR algorithm and facial recognition consists of a 512-dimensional vector generated through AdaFace~\citep{kim2022adaface} with a ResNet backbone~\citep{he2016deep} trained on the CASIA-WebFace dataset~\citep{yi2014learning}. Image descriptors were compared using the cosine similarity metric. For the recognition task, we compare the SR-recovered images against the gallery images. Our proposed algorithm is compared against \gls*{sota} algorithms: SPARNET~\citep{chen2020learning}, GFPGAN~\citep{wang2021gfpgan}, SwinIR~\citep{liang2021swinir}, SDE-SR~\citep{dos2022face}, IDM~\citep{gao2023implicit}, SR3~\citep{saharia2023image}, and SRDG~\citep{visapp24sr}.

\subsection{Training and Evaluation of the \texorpdfstring{$\delta$}{delta} Network}

\label{sub:effectiveness}
\begin{majorblock}
To train the $\delta$ network, we used the CASIA-WebFace dataset~\citep{yi2014learning}. 
In our experiments a total of 490{,}623 images from 10{,}572 identities were used, corresponding to an average of 46.41 images per subject. 
The dataset was divided into training (85\%) and validation (15\%) subsets. 
The model was trained for up to 20 epochs, with early stopping applied when no improvement in the validation loss was observed for five consecutive epochs.

To compute the triplet loss, two high-resolution images were randomly selected for each identity. 
The mean of their feature vectors was used as the positive sample, while the features extracted from their corresponding low-resolution versions were used as inputs to the Feature Combiner module to generate the anchor representation. 
The negative sample was obtained from the high-resolution features of a different identity.

The work of \cite{hermans2017defense} shows that training can collapse when the margin is too large relative to the initial embedding spread, preventing the embeddings from separating properly.
Following this observation, the margin was empirically set slightly above the average gap between the anchor–positive and anchor–negative distances to enhance training stability, resulting in a final value of 0.495.
\end{majorblock}

To demonstrate the effectiveness of the proposed ensemble of FC modules in merging low-resolution features, we conducted experiments on the CelebA dataset using the identities selected in Section~\ref{sec_exp}. For each identity, the features of the low-resolution images \( \text{LR}_1, \dots, \text{LR}_N \) are fused using the ensemble of FCs, and the resulting feature representation is compared to the corresponding ground truth gallery feature via cosine similarity.
We considered two approaches: $(i)$~\(\eta = 0\), where only the arithmetic mean is used for merging low-resolution features, and $(ii)$~\(\eta = 1\), where the network \(\delta\) refines the arithmetic mean during feature integration. This comparison highlights the advantage of using the network \(\delta\) for refinement over a simple averaging scheme.

\begin{figure}[!htb]
    \centering
    \includegraphics[width=0.98\linewidth]{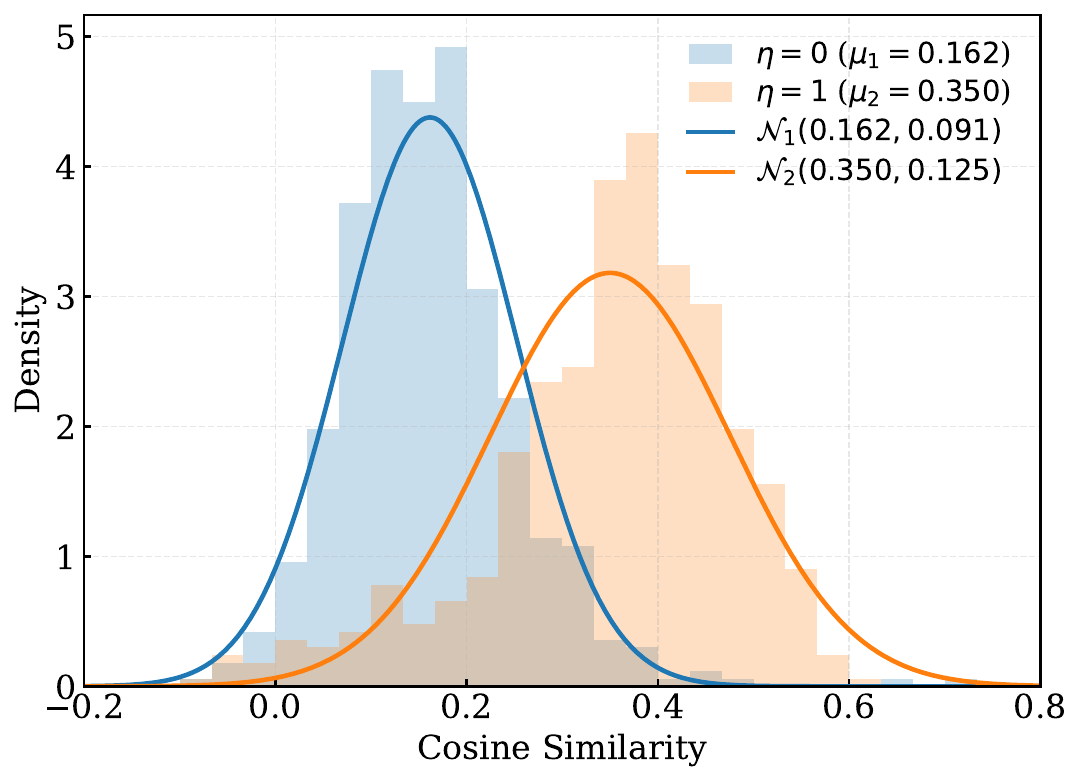}

\caption{\textbf{Histogram of the similarity score distributions for the two approaches: \(\eta=0\) and \(\eta=1\).}
The average similarity score increases substantially from \(\mathbf{0.162}\) when \(\eta = 0\) to \(\mathbf{0.350}\) when \(\eta = 1\).
We also adjusted and plotted two normal distributions $\mathcal{N}(\mu,\sigma)$ that fit the score distributions. A paired t-test yielded \(t = {-}46.33\) and \(p\text{-value} < 10^{\text{-}12}\), demonstrating that the proposed network \(\delta\) considerably enhances feature merging, producing representations that more closely match the high-resolution gallery~features.}

    \label{fig:compare_scores}
\end{figure}

\major{The results show that the mean similarity score increased from 0.162 for \(\eta = 0\) to 0.350 for \(\eta = 1\), representing a 116\% improvement. 
A paired \textit{t}-test confirmed that this difference is highly significant (\(t=-46.33\), \(p<10^{-12}\)), demonstrating that the proposed $\delta$ network effectively enhances feature merging and produces representations that more closely match the high-resolution gallery features.}
The histogram analysis in Figure~\ref{fig:compare_scores} further supports this conclusion, showing a clear rightward shift in the score distribution when the $\delta$ network is applied to refine the merging process. 
This improvement in the feature space ultimately translates into better face recognition performance, as discussed in Section~\ref{sec:general_results}.

\subsection{General Results}
\label{sec:general_results}

In this section, we present the results obtained in our experiments. Tables \ref{quant_celeba} to \ref{tab4} summarize the main findings. Dark gray highlights the best values, while light gray indicates the second-best.

Table \ref{quant_celeba} shows the quantitative results of our proposed method on the CelebA dataset. 
We used the gallery image as a reference for the first three metrics, \acrfull*{auc}, Rank-1, and Rank-5, and the high-resolution ground truth image as a reference for the remaining three metrics: PSNR, SSIM, and \major{LPIPS}.
\major{Compared to other algorithms, FASR++ provides superior results in all metrics, demonstrating its ability to recover discriminative identity features while maintaining high perceptual and structural image quality.}

In general, the parameters of super-resolution algorithms are optimized to improve image quality metrics such as PSNR and SSIM; however, these improvements may be accompanied by distortions in identity and impaired recognition performance.
Moreover, \cite{dos2022face} exemplifies that for a given set of solutions for a super-resolution problem, the images with higher values for PSNR and SSIM metrics are not the images that provide the highest value for recognition metrics. 
Remarkably, FASR++ maintains superior PSNR and SSIM values, \major{and lower LPIPS scores}, while still achieving optimal recognition performance by generating high-quality features through an ensemble of \gls*{fc} modules to assist the diffusion~model.

\begin{table*}[!ht]
\centering
\caption{\textbf{Results for the CelebA Dataset.} 1:1 verification and 1:N identification results, along with SSIM, PSNR, and LPIPS metrics, on both low-resolution (LR) and super-resolved versions.}

\resizebox{0.9\linewidth}{!}{
\begin{tabular}{l c c c c c c}
\toprule
SR Method & \multicolumn{1}{c}{\textbf{AUC}} & \multicolumn{1}{c}{\textbf{Rank-$1$ (\%)}} & \multicolumn{1}{c}{\textbf{Rank-$5$ (\%)}} & \textbf{PSNR}$\uparrow$ & \textbf{SSIM}$\uparrow$ & \textbf{LPIPS}$\downarrow$\\ 
\midrule
LR         & $0.885$                & $11.80$             & $29.60$             & $23.2736\,\pm\,\text{\scriptsize 1.9534}$ & $0.6452\,\pm\,\text{\scriptsize 0.0711}$ & $0.4704\,\pm\,\text{\scriptsize 0.0754}$\\[0.5ex]
GFPGAN     & $0.865$                & $19.80$             & $34.80$             & $23.1720\,\pm\,\text{\scriptsize 1.7734}$ & $0.6545\,\pm\,\text{\scriptsize 0.0688}$ & $0.1915\,\pm\,\text{\scriptsize 0.0714}$\\[0.5ex]
SPARNET    & $0.874$                & $21.80$             & $38.60$             & $21.0168\,\pm\,\text{\scriptsize 2.3292}$ & $0.6121\,\pm\,\text{\scriptsize 0.0814}$ & $0.3014\,\pm\,\text{\scriptsize 0.0869}$\\[0.5ex]
SR3        &$0.936$ & $44.60$          & $62.60$             & $25.4013\,\pm\,\text{\scriptsize 1.9935}$ & $0.7370\,\pm\,\text{\scriptsize 0.0677}$ & \cellcolor{gray!15}$0.1022\,\pm\,\text{\scriptsize 0.0372}$\\[0.5ex]
SwinIR     & $0.921$                & $40.40$             & $57.20$             & $21.7619\,\pm\,\text{\scriptsize 2.0440}$ & $0.6660\,\pm\,\text{\scriptsize 0.0848}$ & $0.2008\,\pm\,\text{\scriptsize 0.0702}$\\[0.5ex]
SDE-SR      & $0.933$                & $45.60$ & $66.00$ & $25.6987\,\pm\,\text{\scriptsize 2.0236}$ & $0.7522\,\pm\,\text{\scriptsize 0.0659}$ & $0.1116\,\pm\,\text{\scriptsize 0.0392}$\\[0.5ex]
FASR      & \cellcolor{gray!15}$0.946$                & \cellcolor{gray!15}$53.20$ & \cellcolor{gray!15}$68.60$ & \cellcolor{gray!15}$25.8966\,\pm\,\text{\scriptsize 2.0739}$ & \cellcolor{gray!15}$0.7588\,\pm\,\text{\scriptsize 0.0657}$ & $0.1191\,\pm\,\text{\scriptsize 0.0427}$\\[0.5ex]
\textbf{FASR++ (Ours)}& \cellcolor{gray!30}$0.951$ & \cellcolor{gray!30}$59.60$ & \cellcolor{gray!30}$74.80$ & \cellcolor{gray!30}$26.6413\,\pm\,\text{\scriptsize 2.1197}$ & \cellcolor{gray!30}$0.7756\,\pm\,\text{\scriptsize 0.0643}$ & \cellcolor{gray!30}$0.0983\,\pm\,\text{\scriptsize 0.0373}$\\
\bottomrule
\end{tabular}
}
\label{quant_celeba}
\end{table*}

\begin{table*}[!ht]
\centering
\caption{\textbf{Results for the Quis-Campi Dataset.} 1:1 verification and 1:N identification results, along with SSIM, PSNR, and LPIPS metrics, on both low-resolution (LR) and super-resolved versions.}
\resizebox{0.9\linewidth}{!}{
\begin{tabular}{l c c c c c c}
\toprule
SR Method & \multicolumn{1}{c}{\textbf{AUC}} & \multicolumn{1}{c}{\textbf{Rank-$1$ (\%)}} & \multicolumn{1}{c}{\textbf{Rank-$5$ (\%)}} & \textbf{PSNR}$\uparrow$ & \textbf{SSIM}$\uparrow$ & \textbf{LPIPS}$\downarrow$\\ 
\midrule
LR         & $0.815$                & $31.11$             & $51.78$             & $29.1295\,\pm\,\text{\scriptsize 3.0910}$ & $0.8257\,\pm\,\text{\scriptsize 0.0612}$ & $0.3150\,\pm\,\text{\scriptsize 0.0814}$\\[0.5ex]
GFPGAN     & $0.789$                & $17.78$             & $42.22$             & $27.8496\,\pm\,\text{\scriptsize 2.6545}$ & $0.7733\,\pm\,\text{\scriptsize 0.0640}$ & $0.2575\,\pm\,\text{\scriptsize 0.0557}$\\[0.5ex]
SPARNET    & $0.862$                & $32.67$             & $58.44$             & $24.9380\,\pm\,\text{\scriptsize 3.9794}$ & $0.7663\,\pm\,\text{\scriptsize 0.0986}$ & $0.2092\,\pm\,\text{\scriptsize 0.0794}$\\[0.5ex]
SR3        & $0.914$                & $46.00$             & $70.89$             & $30.7066\,\pm\,\text{\scriptsize 3.0262}$ & $0.8477\,\pm\,\text{\scriptsize 0.0558}$ & \cellcolor{gray!15}$0.1235\,\pm\,\text{\scriptsize 0.0372}$\\[0.5ex]
SRDG       & \cellcolor{gray!30}$0.920$   & $46.89$             & \cellcolor{gray!15}$73.33$             & $30.0598\,\pm\,\text{\scriptsize 2.8980}$ & $0.8176\,\pm\,\text{\scriptsize 0.0576}$ & $0.1410\,\pm\,\text{\scriptsize 0.0459}$\\[0.5ex]
IDM        & $0.884$                & $30.89$             & $59.11$             & $26.2468\,\pm\,\text{\scriptsize 3.6993}$ & $0.7522\,\pm\,\text{\scriptsize 0.0866}$ & $0.1663\,\pm\,\text{\scriptsize 0.0501}$\\[0.5ex]
SDE-SR      & $0.916$   & $47.11$             & $71.56$             & $30.3447\,\pm\,\text{\scriptsize 2.8344}$ & $0.8250\,\pm\,\text{\scriptsize 0.0546}$ & $0.1298\,\pm\,\text{\scriptsize 0.0374}$\\[0.5ex]
FASR      & $0.917$   & \cellcolor{gray!15}$50.67$             & $72.22$             & \cellcolor{gray!15}$30.7092\,\pm\,\text{\scriptsize 2.7705}$ & \cellcolor{gray!15}$0.8484\,\pm\,\text{\scriptsize 0.0496}$ & $0.1303\,\pm\,\text{\scriptsize 0.0342}$\\[0.5ex]
\textbf{FASR++ (Ours)} & \cellcolor{gray!15}$0.918$   & \cellcolor{gray!30}$52.22$             & \cellcolor{gray!30}$75.11$             & \cellcolor{gray!30}$31.7817\,\pm\,\text{\scriptsize 2.9570}$ & \cellcolor{gray!30}$0.8554\,\pm\,\text{\scriptsize 0.0491}$ & \cellcolor{gray!30}$0.0983\,\pm\,\text{\scriptsize 0.0309}$\\
\bottomrule
\end{tabular}
}
\label{tab2}
\end{table*}

Table~\ref{tab2} presents the quantitative results on the Quis-Campi dataset.
The SRDG algorithm uses soft attributes as input to guide its diffusion process during image reconstruction.
Although SRDG benefits from this additional conditioning, our method surpasses SRDG and the other competing algorithms in both recognition metrics (Rank-1 and Rank-5) and image quality metrics (PSNR, SSIM, and \major{LPIPS}).%

\major{
To further assess the reliability of the image quality improvements, a paired \textit{t}-test was conducted to evaluate the statistical significance of the results across both datasets. 
FASR++ was compared against the second-best performing methods: FASR for PSNR and SSIM, and SR3 for LPIPS. 
On the CelebA dataset, the improvements in PSNR and SSIM were confirmed to be statistically significant relative to FASR ($t_\text{PSNR}=28.9$, $t_\text{SSIM}=22.4$), while the difference in perceptual quality, measured by LPIPS, was also statistically significant compared to SR3 ($t_\text{LPIPS}=-4.4$). 
Similarly, on the Quis-Campi dataset, statistically significant differences were observed when comparing FASR++ to FASR for PSNR and SSIM ($t_\text{PSNR}=26.4$, $t_\text{SSIM}=8.9$), and to SR3 for LPIPS ($t_\text{LPIPS}=-16.1$). 
In all cases, the differences were highly significant ($p<10^{-5}$), confirming that the observed gains are not due to random variation but instead establish the proposed method as the \gls*{sota} in both reconstruction fidelity and perceptual quality.}

\begin{figure*}[!htb]
\centering
\setlength{\tabcolsep}{1pt}
\resizebox{0.9\linewidth}{!}{
\begin{tabular}{cccccc}
\miniscule{LR} & \miniscule{GFPGAN} & \miniscule{SR3} & \miniscule{SDE-SR} & \miniscule{\textbf{FASR++ (Ours)}} & \miniscule{GT}\\[-0.25ex]

\includegraphics[width=10mm]{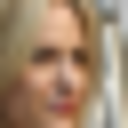} &
\includegraphics[width=10mm]{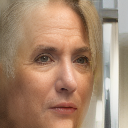} &
\includegraphics[width=10mm]{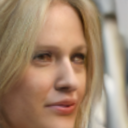} &
\includegraphics[width=10mm]{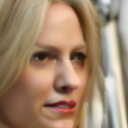} &
\includegraphics[width=10mm]{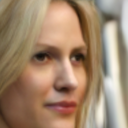} &
\includegraphics[width=10mm]{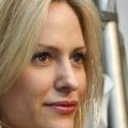} \\[-0.4ex]

\includegraphics[width=10mm]{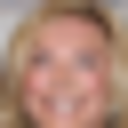} &
\includegraphics[width=10mm]{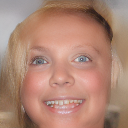} &
\includegraphics[width=10mm]{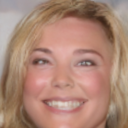} &
\includegraphics[width=10mm]{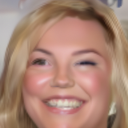} &
\includegraphics[width=10mm]{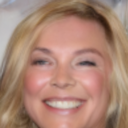} &
\includegraphics[width=10mm]{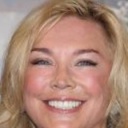} \\[-0.4ex]

\includegraphics[width=10mm]{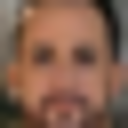} &
\includegraphics[width=10mm]{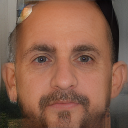} &
\includegraphics[width=10mm]{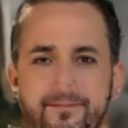} &
\includegraphics[width=10mm]{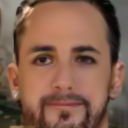} &
\includegraphics[width=10mm]{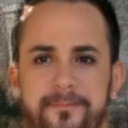} &
\includegraphics[width=10mm]{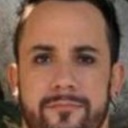} \\[-0.4ex]

\includegraphics[width=10mm]{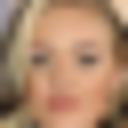} &
\includegraphics[width=10mm]{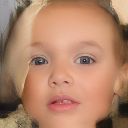} &
\includegraphics[width=10mm]{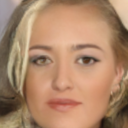} &
\includegraphics[width=10mm]{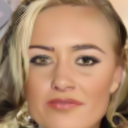} &
\includegraphics[width=10mm]{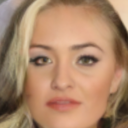} &
\includegraphics[width=10mm]{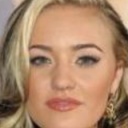} \\[-0.4ex]

\includegraphics[width=10mm]{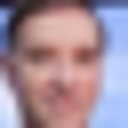} &
\includegraphics[width=10mm]{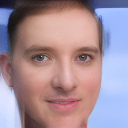} &
\includegraphics[width=10mm]{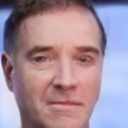} &
\includegraphics[width=10mm]{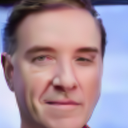} &
\includegraphics[width=10mm]{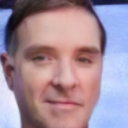} &
\includegraphics[width=10mm]{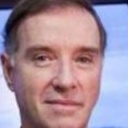} \\[-0.4ex]

\end{tabular}
} 
\caption{\textbf{Qualitative results for the CelebA Dataset.} Comparison of low-resolution~(LR) images, super-resolution~(SR) outputs obtained by various methods, and ground truth~(GT) images. \gls*{srname} outperforms the baselines by preserving facial symmetry and ensuring a natural~appearance.}

\label{fig_quali_celeba}
\end{figure*}

\begin{figure*}[!htb]
\centering
\setlength{\tabcolsep}{1pt}
\resizebox{0.9\linewidth}{!}{ %
\begin{tabular}{ccccccc}
\miniscule{LR}&\miniscule{SR3}&\miniscule{IDM}&\miniscule{SDE-SR}&\miniscule{SRDG}&\miniscule{\textbf{FASR++ (Ours)}}&\miniscule{GT}\\[-0.25ex]

 \includegraphics[width=10mm]{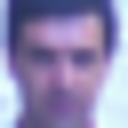} &  \includegraphics[width=10mm]{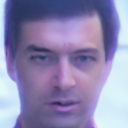}&  \includegraphics[width=10mm]{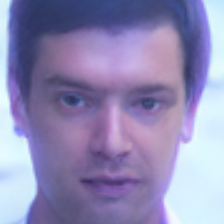}&\includegraphics[width=10mm]{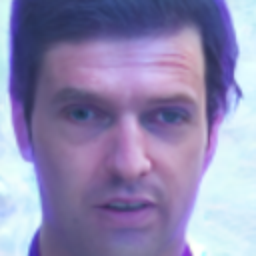}   &  \includegraphics[width=10mm]{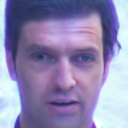} &  \includegraphics[width=10mm]{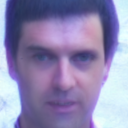}&\includegraphics[width=10mm]{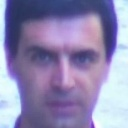}  \\[-0.4ex]
 \includegraphics[width=10mm]{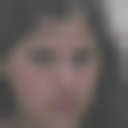} &  \includegraphics[width=10mm]{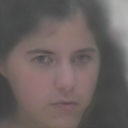}& \includegraphics[width=10mm]{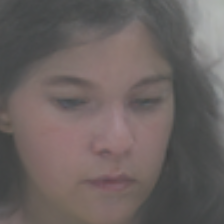}&\includegraphics[width=10mm]{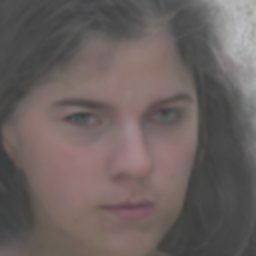}   &  \includegraphics[width=10mm]{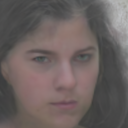} &  \includegraphics[width=10mm]{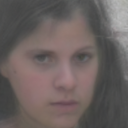}&\includegraphics[width=10mm]{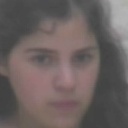}  \\[-0.4ex]
 \includegraphics[width=10mm]{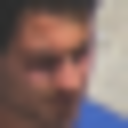} &  \includegraphics[width=10mm]{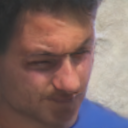}&  \includegraphics[width=10mm]{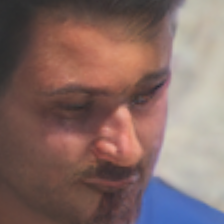}&\includegraphics[width=10mm]{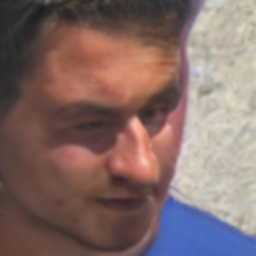}   &  \includegraphics[width=10mm]{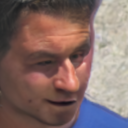} &  \includegraphics[width=10mm]{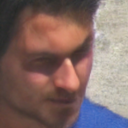}&\includegraphics[width=10mm]{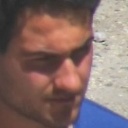}  \\[-0.4ex]
 \includegraphics[width=10mm]{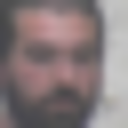} &  \includegraphics[width=10mm]{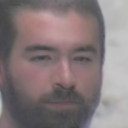}&  \includegraphics[width=10mm]{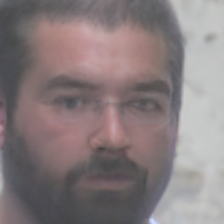}&\includegraphics[width=10mm]{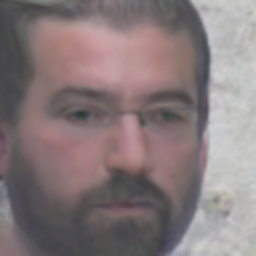}   &  \includegraphics[width=10mm]{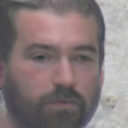} &  \includegraphics[width=10mm]{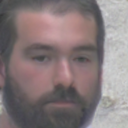} &\includegraphics[width=10mm]{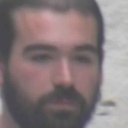} \\[-0.4ex]
\end{tabular}
} 
\caption{\textbf{Qualitative Results for the Quis-Campi Dataset.} Comparison of low-resolution~(LR) images, super-resolution~(SR) outputs obtained by various methods, and ground truth~(GT) images. \gls*{srname} outperforms the baselines by preserving facial symmetry and ensuring a natural~appearance.}

\label{fig_quali}
\end{figure*}

In Figures \ref{fig_quali_celeba} and \ref{fig_quali}, we present the qualitative comparison of our method \gls*{srname} against other super-resolution algorithms for the Celeba and Quis-Campi datasets, respectively.
While all other methods are effective to some extent, they often introduce artifacts or noise into the facial images, typical issues encountered in \gls*{sr} algorithms. For instance, in most examples, the images generated by other algorithms exhibit distortions, mainly in the eye region, providing an artificial and distorted appearance. In Figure \ref{fig_quali_celeba}, GFPGAN yields distortions regarding the person's age.
In contrast, \gls*{srname} stands out as the only approach that produces natural-looking images without noticeable artificiality. It preserves symmetries and successfully recovers details without introducing artifacts or distorting facial~features. %

Due to the ill-posed nature of the SR problem, many SR algorithms suffer from bias issues and struggle to recover a person's identity accurately. In contrast, our algorithm effectively tackles these challenges, mitigating identity-related problems and yielding superior quantitative and qualitative~results.

\subsection{Ablation Study}
\begin{figure*}[!ht]
    \centering
    \begin{minipage}{0.45\linewidth}
        \centering
        \includegraphics[width=\linewidth]{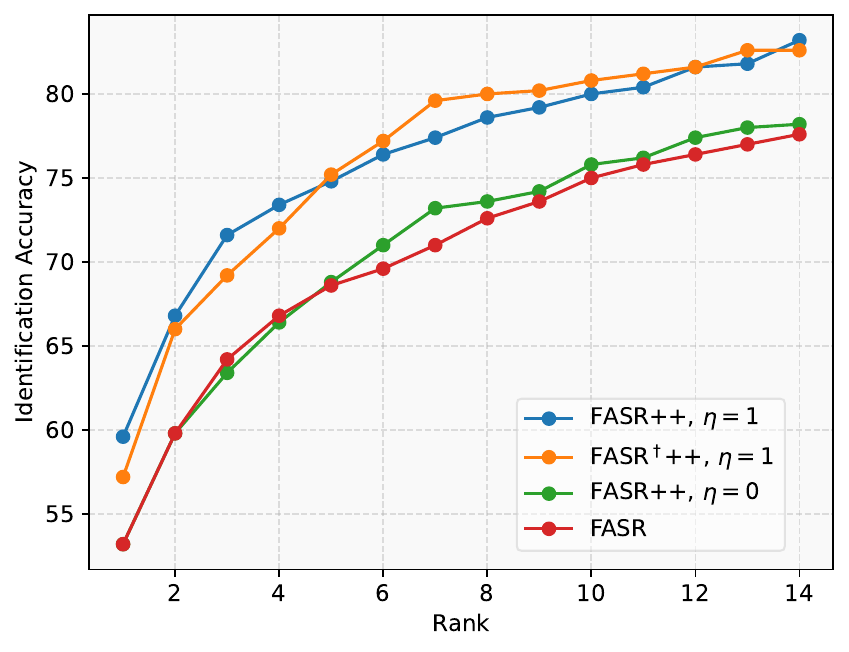}
    \end{minipage}
    \qquad
    \begin{minipage}{0.45\linewidth}
        \centering
        \includegraphics[width=\linewidth]{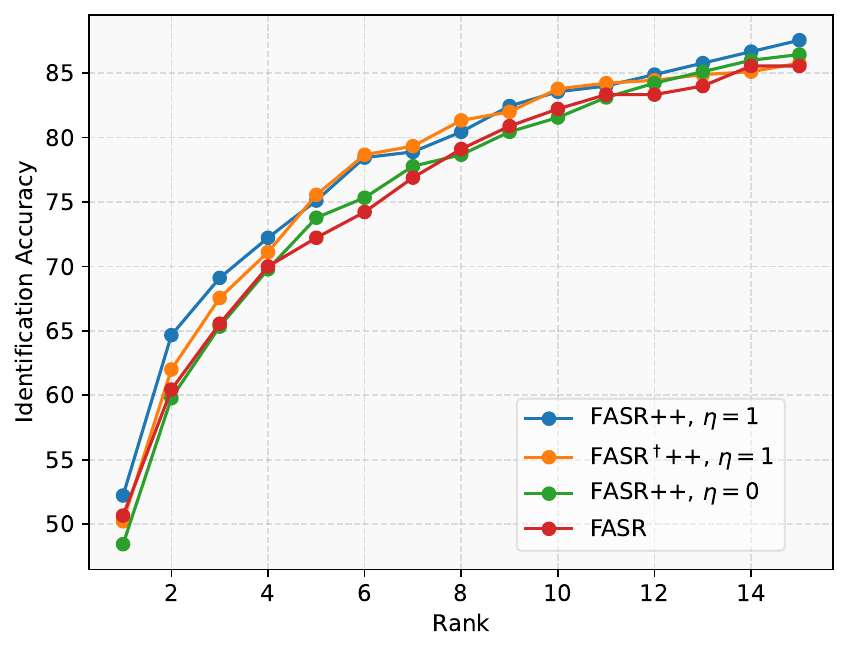}
    \end{minipage}
\caption{\textbf{CMC curves on CelebA~(left) and Quis-Campi~(right) datasets.} 
The evaluation includes four super-resolution methods: FASR, FASR++ with \(\eta = 0\) and \(\eta = 1\), and FASR$^\dagger$++ (where the \(\delta\) network was trained on a subset of the CelebA dataset). 
The proposed FASR++ improves recognition accuracy across both datasets, demonstrating its effectiveness over the baseline methods.}

    \label{fig:cmc_curves}
\end{figure*}

\begin{majorblock}
This section analyzes the impact of the $\delta$ network and the choice of dataset on both recognition accuracy and image quality, followed by an evaluation of how the number of training samples used in the $\delta$ network influences its effectiveness and the overall performance of the model.

\paragraph{Influence of the $\delta$ Network and Dataset.}
In this section, we conduct an incremental study to examine the impact of each enhancement applied to our proposed method on the CelebA and Quis-Campi datasets. We start with \textbf{FASR}, where low-resolution features are combined using an arithmetic mean to produce a single super-resolution output. Next, we introduce \textbf{FASR++ (\(\eta=0\))}, which still combines low-resolution features via arithmetic mean but generates two super-resolution images that are averaged to create a final image. 
We also present two versions of our complete approach, \textbf{FASR++} and \textbf{FASR$^\dagger$++}. 
In both cases, an ensemble of Feature Combiner modules integrated with the $\delta$ network ($\eta=1$) is employed to estimate a reliable descriptor and to produce two super-resolution images. 
As described in Section~\ref{sub:effectiveness}, in \textbf{FASR++} the $\delta$ network is trained on the CASIA-WebFace dataset using 10{,}572 identities, with an average of 46.41 images per subject. 
Alternatively, in \textbf{FASR$^\dagger$++}, the $\delta$ network is trained on a subset of the CelebA dataset, where we selected 2{,}000 identities with an average of 20 images per identity. 
This experiment aims to analyze the influence of dataset limitations such as fewer identities, reduced intra-class variability, and a smaller number of images per identity on the performance of the proposed fusion mechanism.

To further examine the role of the weighting factor $\eta$ in the fusion process, we performed additional experiments with intermediate values $\eta \in \{0, 0.25, 0.5, 0.75, 1.0\}$ using the same experimental setup described in Section~\ref{sub:effectiveness}. For $\eta \neq 0$, the resulting cosine similarity distributions were largely overlapping, indicating that $\eta$ has little effect on the fused representations. This behavior occurs because the FC module is trained to optimize the combined feature representation as a whole, so any scaling change in $\eta$ tends to be internally compensated by the learned weights of the $\delta$ network, maintaining consistent FC outputs across different configurations. Therefore, $\eta = 1$ was adopted in all experiments.

Tables~\ref{tab3} and~\ref{tab4} present the verification and recognition results along with the corresponding image quality metrics. 
In addition, the detailed CMC curves for the CelebA and Quis-Campi datasets are shown in Figure~\ref{fig:cmc_curves}.
From Table~\ref{tab3}, we observe a consistent improvement as each component is introduced in our incremental study. 
Averaging two images in {FASR++} ($\eta=0$) slightly enhances reconstruction quality compared to the baseline {FASR}, while maintaining similar recognition performance. 
When the learned fusion mechanism is activated ($\eta=1$) in {FASR$^\dagger$++}, both recognition accuracy and perceptual quality improve, with a more pronounced effect on recognition (a $4\%$ increase in Rank-1 accuracy). 
Finally, training $\delta$ on the larger CASIA-WebFace dataset ({FASR++}) yields the best overall Rank-1 accuracy (a $6.4\%$ improvement over the baseline), whereas for the other metrics, both {FASR$^\dagger$++} and {FASR++} achieve very similar~results.

\begin{table*}[!htb]
\centering
\caption{\textbf{Ablation Study on the CelebA dataset.} We evaluate the impact of our FASR++ approach by comparing three configurations: FASR, FASR++ with $\eta=0$, and FASR$^\dagger$++ with $\eta=1$. In the latter case, the $\delta$ network was trained on a subset of the CelebA dataset.}

\resizebox{0.9\linewidth}{!}{
\begin{tabular}{l c c c c c c c}
\toprule
\textbf{SR Method} & $\eta$ & \textbf{AUC} & \textbf{Rank-1 (\%)} & \textbf{Rank-5 (\%)} & \textbf{PSNR}$\uparrow$ & \textbf{SSIM}$\uparrow$ & \textbf{LPIPS}$\downarrow$ \\ 
\midrule
FASR              & 0 & 0.946 & 53.20 & 68.60 & $25.8966\,\pm\,\text{\scriptsize 2.0739}$ & $0.7588\,\pm\,\text{\scriptsize 0.0657}$ & $0.1191\,\pm\,\text{\scriptsize 0.0427}$ \\[0.5ex]
FASR++            & 0 & 0.944 & 53.20 & 68.80 &$26.5421\,\pm\,\text{\scriptsize 2.0690}$ &$0.7701\,\pm\,\text{\scriptsize 0.0637}$ &$0.0989\,\pm\,\text{\scriptsize 0.0362}$ \\[0.5ex]
FASR$^\dagger$++  & 1 & \cellcolor{gray!15}0.950 & \cellcolor{gray!15}57.20 & \cellcolor{gray!30}75.20 &\cellcolor{gray!15}$26.5531\,\pm\,\text{\scriptsize 2.0815}$ & \cellcolor{gray!15}$0.7719\,\pm\,\text{\scriptsize 0.0641}$ & \cellcolor{gray!30}$0.0938\,\pm\,\text{\scriptsize 0.0351}$ \\[0.5ex]
\textbf{FASR++}   & 1 & \cellcolor{gray!30}0.951 & \cellcolor{gray!30}59.60 & \cellcolor{gray!15}74.80 & \cellcolor{gray!30}$26.6413\,\pm\,\text{\scriptsize 2.1197}$ & \cellcolor{gray!30}$0.7756\,\pm\,\text{\scriptsize 0.0643}$ & \cellcolor{gray!15}$0.0983\,\pm\,\text{\scriptsize 0.0373}$ \\
\bottomrule
\end{tabular}
}
\label{tab3}
\end{table*}

\begin{table*}[!htb]
\centering
\caption{\textbf{Ablation Study on the Quis-Campi dataset.} We evaluate the impact of our FASR++ approach by comparing three configurations: FASR, FASR++ with $\eta=0$, and FASR$^\dagger$++ with $\eta=1$. In the latter case, the $\delta$ network was trained on a subset of the CelebA dataset.}
\resizebox{0.9\linewidth}{!}{
\begin{tabular}{l c c c c c c c}
\toprule
\textbf{SR Method} & $\eta$ & \textbf{AUC} & \textbf{Rank-1 (\%)} & \textbf{Rank-5 (\%)} & \textbf{PSNR}$\uparrow$ & \textbf{SSIM}$\uparrow$ & \textbf{LPIPS}$\downarrow$\\ 
\midrule
FASR              & 0 & \cellcolor{gray!15}0.917 & \cellcolor{gray!15}50.67 & 72.22 & $30.7092\,\pm\,\text{\scriptsize 2.7705}$ & $0.8484\,\pm\,\text{\scriptsize 0.0496}$ & $0.1303\,\pm\,\text{\scriptsize 0.0342}$ \\[0.5ex]
FASR++            & 0 & 0.915 & 48.44 & 73.78 & \cellcolor{gray!15}$31.7033\,\pm\,\text{\scriptsize 2.9282}$ & \cellcolor{gray!15}$0.8531\,\pm\,\text{\scriptsize 0.0487}$ & $0.1004\,\pm\,\text{\scriptsize 0.0310}$ \\[0.5ex]
FASR$^\dagger$++  & 1 & \cellcolor{gray!30}0.918 &50.22 & \cellcolor{gray!30}75.56 & $31.6540\,\pm\,\text{\scriptsize 2.9476}$ & $0.8516\,\pm\,\text{\scriptsize 0.0511}$ & \cellcolor{gray!15}$0.1011\,\pm\,\text{\scriptsize 0.0319}$ \\[0.5ex]
\textbf{FASR++}   & 1 & \cellcolor{gray!30}0.918 & \cellcolor{gray!30}52.22 & \cellcolor{gray!15}75.11 & \cellcolor{gray!30}$31.7817\,\pm\,\text{\scriptsize 2.9570}$ & \cellcolor{gray!30}$0.8554\,\pm\,\text{\scriptsize 0.0491}$ & \cellcolor{gray!30}$0.0983\,\pm\,\text{\scriptsize 0.0309}$ \\
\bottomrule
\end{tabular}
}
\label{tab4}
\end{table*}

From Table~\ref{tab4}, we observe a similar incremental behavior on the Quis-Campi dataset. 
Averaging two images in {FASR++} ($\eta=0$) results in a clear perceptual improvement, yielding a substantial reduction in LPIPS (from $0.1303$ to $0.1004$) compared to FASR.
When the learned fusion mechanism is enabled ($\eta=1$) in {FASR$^\dagger$++}, the gain is more pronounced in Rank-5 accuracy. 
Finally, training $\delta$ on the larger CASIA-WebFace dataset ({FASR++}) yields the best overall results, except for Rank-5 accuracy.

For both datasets, we observe that using the $\delta$ network ($\eta=1$) provides superior results compared to the other configurations. 
This outcome was expected since, as discussed in Section~\ref{sub:effectiveness}, the $\delta$ network combines features efficiently. 
The improved results in both recognition and image quality metrics confirm that the combined features are being effectively leveraged by the diffusion model. 
Moreover, the larger number of images and the greater identity diversity of the CASIA-WebFace dataset contribute to the superior results of {FASR++} compared to {FASR$^\dagger$++}, mainly in terms of Rank-1~accuracy.
\end{majorblock}

We can also observe that the superior results of {FASR++} are more pronounced for the CelebA dataset and more subtle for Quis-Campi. This can be attributed to two main reasons: $(i)$~as will be described in the next section, the Quis-Campi dataset is a real-world surveillance dataset, which is more challenging, containing variations in lighting and pose, as well as a higher level of noise compared to CelebA. Consequently, the facial features in Quis-Campi can be noisier and more difficult to reconstruct accurately; $(ii)$~for the Quis-Campi dataset, only five features from low-resolution images were used to assist in the reconstruction of the super-resolution images, while more than eight  low-resolution images were used for CelebA, providing a richer source of information for the algorithm.

\begin{figure}[!ht]
    \centering
    \begin{minipage}{0.99\linewidth}
        \centering
        \includegraphics[width=\linewidth]{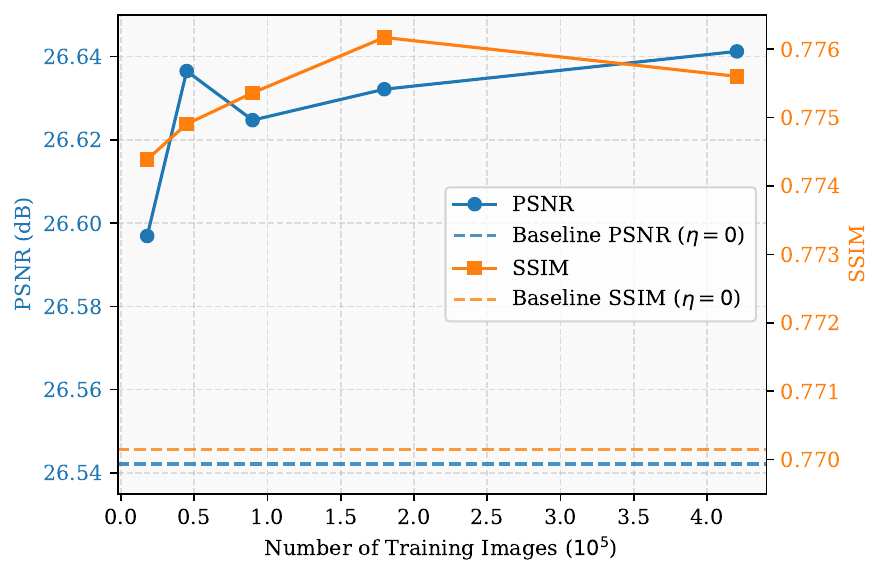}
    \end{minipage}\\[0.4em] %

    \begin{minipage}{0.821\linewidth}
        \centering
        \hspace{-0.6cm}\includegraphics[width=\linewidth]{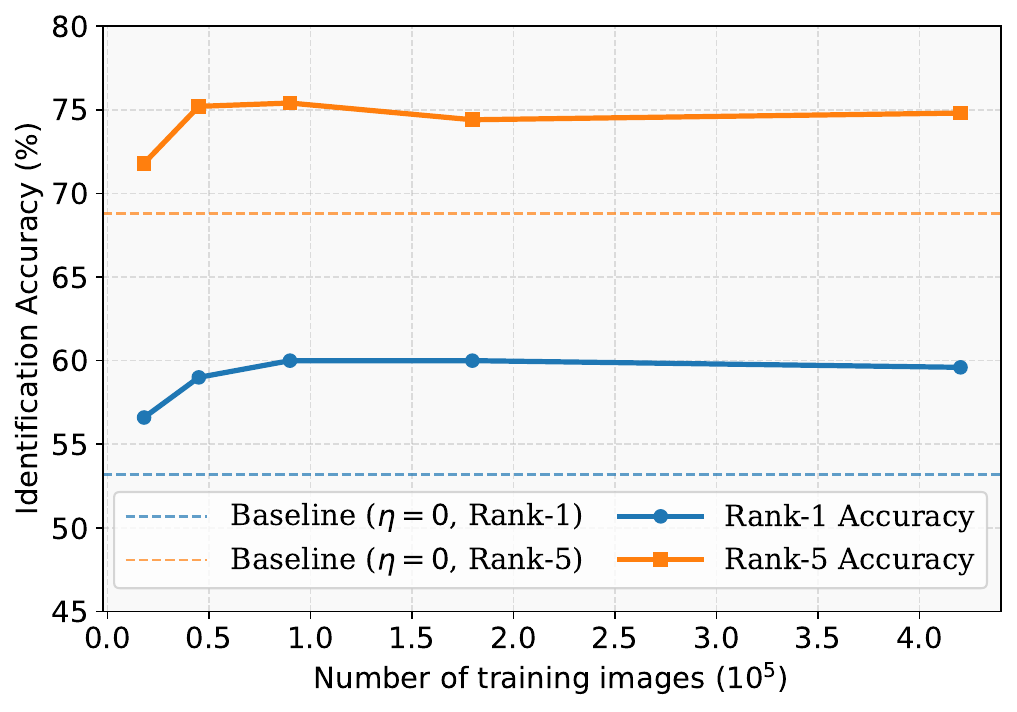}
    \end{minipage}
\caption{
Performance metrics of FASR++ as a function of the number of training images.  
The top plot illustrates the image quality metrics (PSNR and SSIM), whereas the bottom plot presents the recognition metrics (Rank-1 and Rank-5 accuracies).
}
    \label{fig:metrics_numimgs}
\end{figure}

\major{
\paragraph{Effect of the Training Dataset Size on the $\delta$ Network.}
We now analyze how the total number of samples used for training the $\delta$ network affects its ability to effectively merge features and influence the final results of the FASR++ framework, using the CelebA dataset as the evaluation benchmark.
In this experiment, we limit the number of images per identity in the CASIA-WebFace dataset and evaluate the impact on the PSNR, SSIM, Rank-1, and Rank-5 metrics. 
Figure~\ref{fig:metrics_numimgs} shows that as the number of training samples increases, 
both image quality metrics (PSNR and SSIM) and recognition metrics (Rank-1 and Rank-5 accuracies) 
remain relatively stable with only minor fluctuations, suggesting that approximately 50k images are sufficient to achieve optimal performance in both reconstruction quality and face recognition. 
In this range, the Rank-1 and Rank-5 scores converge to around 60\% and 75\%, respectively, 
while statistical analysis confirmed that, despite small variations in PSNR and SSIM, 
no significant differences were observed when training with 50k images or more ($p > 0.05$).}

\subsection{Failure Cases}

\begin{figure}[!htb]
\centering
\setlength{\tabcolsep}{1.2pt}
\resizebox{0.8\linewidth}{!}{ %
\begin{tabular}{ccc}
\tiny{SRDG}&\tiny{\textbf{FASR++ (Ours)}} &\tiny{GT}\\[-0.3ex]
\includegraphics[width=15mm]{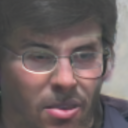} & 
\includegraphics[width=15mm]{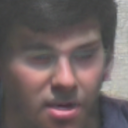} &\includegraphics[width=15mm]{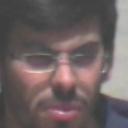} \\[-0.25ex]

\tiny{SDE-SR}&\tiny{\textbf{FASR++ (Ours)}} &\tiny{GT}\\[-0.3ex]
\includegraphics[width=15mm]{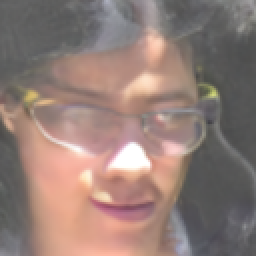} &
\includegraphics[width=15mm]{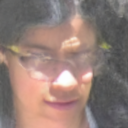} & 
\includegraphics[width=15mm]{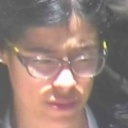} \\
\end{tabular}
}
\caption{\textbf{Failure cases.} The first row presents results from SRDG~\citep{visapp24sr}, \gls*{srname}~(ours), and ground truth~(GT) images, while the second row presents results from SDE-SR~\citep{dos2022face}, \gls*{srname}~(ours), and GT images.}
\vspace{3mm}
\label{fig_fail}
\end{figure}

Figure \ref{fig_fail} shows some failure cases of our algorithm compared to SRDG and SDE-SR.
In the first row, \gls*{srname} fails to recover the eyeglasses correctly, whereas SRDG successfully recovers this attribute.
However, it is important to note that SRDG requires explicit information on whether the person is wearing eyeglasses.
This information is not always discernible from \gls*{lr} images in surveillance~scenarios.

In the second row of Figure~\ref{fig_fail}, we observe a failure case of \gls*{srname} compared to SDE-SR.
The image in question shows significant pose variation and highly heterogeneous illumination.
\gls*{srname} produces smoother images with less noise than the other algorithms, causing the information about eyeglasses and the sun's reflection to spread across the periocular~region.

Upon closer examination of the cases where our algorithm fails in Rank-5, we observed that most images share characteristics similar to those described in the previous paragraphs.
Thus, \gls*{srname} provides better results for recognition accuracy but may be more sensitive to variations in pose and~lighting.

\section{Conclusions}
\label{sec:Conclusions}
In this work, we introduced \gls*{srname}, an algorithm that effectively combines multiple features through a neural network to produce a reliable and representative feature vector.
This vector is then integrated with a reference low-resolution image in a diffusion model to generate high-quality super-resolution images.
A key advantage of our algorithm is its independence from explicitly provided facial attributes; instead, it implicitly extracts high-level information through a visual encoder.
This methodology enables our algorithm to preserve individuals' identities more effectively than other methods, resulting in high-quality SR images with enhanced face symmetry, reduced noise and minimized distortion of facial attributes.
We validated our approach on the CelebA and Quis-Campi datasets and achieved state-of-the-art results for visual quality and recognition metrics, demonstrating its potential for applications in real-world surveillance scenarios.

\section*{Declarations}

\begin{contributions}
Marcelo dos Santos is the main contributor and writer of this manuscript. Rayson Laroca assisted in reviewing and editing the manuscript. João Carlos Raposo Neves co-supervised the project and reviewed the manuscript. David Menotti supervised the project, provided resources, and contributed to the review and editing. All authors read and approved the final manuscript.

\end{contributions}

\begin{interests}
The authors declare that they have no competing interests.
\end{interests}

\begin{funding}
    This study was financed by the \textit{Coordena{\c c}{\~a}o de Aperfei{\c c}oamento de Pessoal de N{\'i}vel Superior -- Brasil~(CAPES)} -- Finance Code~001, by the ``V2IP: Videomonitoramento para Identifica{\c c}{\~a}o de Pessoas e Ve{\'i}culos'' CAPES-PROCAD project (\#~88887.619562/2021-00), and by the \textit{Conselho Nacional de Desenvolvimento Cient{\'i}fico e Tecnol{\'o}gico~(CNPq)} (\#~315409/2023-1).
\end{funding}

\begin{materials}

Our code is publicly available at~\textbf{\url{https://github.com/marcelowds/fasrpp}}.
\end{materials}

\bibliographystyle{apalike-sol}
\bibliography{refs}

@INPROCEEDINGS{emil2011,
  author={Bilgazyev, Emil and Efraty, Boris and Shah, Shishir K. and Kakadiaris, Ioannis A.},
  booktitle={International Joint Conference on Biometrics (IJCB)}, 
  title={Improved face recognition using super-resolution}, 
  year={2011},
  volume={},
  number={},
  pages={1-7},
  doi={10.1109/IJCB.2011.6117554}}

@inproceedings{yu2018super,
	title = {Super-Resolving Very Low-Resolution Face Images with Supplementary Attributes},
	author = {Yu, Xin and Fernando, Basura and Hartley, Richard and Porikli, Fatih},
	year = {2018},
	booktitle = {IEEE/CVF Conf. on Computer Vision and Pattern Recognition},
	volume = {},
	number = {},
	pages = {908-917},
	doi = {10.1109/CVPR.2018.00101}
}

@INPROCEEDINGS{liu2015faceattributes,
  author={Liu, Ziwei and Luo, Ping and Wang, Xiaogang and Tang, Xiaoou},
  booktitle={IEEE International Conference on Computer Vision (ICCV)}, 
  title={Deep Learning Face Attributes in the Wild}, 
  year={2015},
  volume={},
  number={},
  pages={3730-3738},
  doi={10.1109/ICCV.2015.425}}

@article{jiang2021deep,
  title={Deep learning-based face super-resolution: A survey},
  author={Jiang, Junjun and Wang, Chenyang and Liu, Xianming and Ma, Jiayi},
  journal={ACM Computing Surveys (CSUR)},
  volume={55},
  number={1},
  pages={1--36},
  year={2021},
  publisher={ACM New York, NY},
doi={10.1145/3485132}
}

@article{saharia2023image,
  title = {Image Super-Resolution via Iterative Refinement},
  author = {Saharia, Chitwan and Ho, Jonathan and Chan, William and Salimans, Tim and Fleet, David J. and Norouzi, Mohammad},
  year = {2023},
  journal = {IEEE Transactions on Pattern Analysis and Machine Intelligence},
  volume = {45},
  number = {4},
  pages = {4713-4726},
  doi = {10.1109/TPAMI.2022.3204461},
}

@article{chen2020learning,
  title={Learning spatial attention for face super-resolution},
  author={Chen, Chaofeng and Gong, Dihong and Wang, Hao and Li, Zhifeng and Wong, Kwan-Yee K},
  journal={IEEE Transactions on Image Processing},
  volume={30},
  pages={1219-1231},
  year={2020},
doi={10.1109/TIP.2020.3043093
}
}

@inproceedings{lu2018attribute,
  title={Attribute-guided face generation using conditional cyclegan},
  author={Lu, Yongyi and Tai, Yu-Wing and Tang, Chi-Keung},
  booktitle={European Conference on Computer Vision (ECCV)},
  pages={282--297},
  year={2018},
doi={10.1007/978-3-030-01258-8\_18}
}

@inproceedings{lee2018attribute,
  title={Attribute augmented convolutional neural network for face hallucination},
  author={Lee, Cheng-Han and Zhang, Kaipeng and Lee, Hu-Cheng and Cheng, Chia-Wen and Hsu, Winston},
  booktitle={IEEE Conference on Computer Vision and Pattern Recognition workshops},
  pages={721--729},
  year={2018},
doi={10.1109/CVPRW.2018.00115}
}

@INPROCEEDINGS{liang2021swinir,
  author={Liang, Jingyun and Cao, Jiezhang and Sun, Guolei and Zhang, Kai and Van Gool, Luc and Timofte, Radu},
  booktitle={2021 IEEE/CVF International Conference on Computer Vision Workshops (ICCVW)}, 
  title={SwinIR: Image Restoration Using Swin Transformer}, 
  year={2021},
  volume={},
  number={},
  pages={1833-1844},
  doi={10.1109/ICCVW54120.2021.00210}}

@inproceedings{santos2024multi,
  title = {Multi-Feature Aggregation in Diffusion Models for Enhanced Face Super-Resolution},
  author = {M. {dos Santos} and R. {Laroca} and R. O. {Ribeiro} and J. {Neves} and D. {Menotti}},
  year = {2024},
  month = {Sept},
  booktitle = {Conference on Graphics, Patterns and Images (SIBGRAPI)},
  volume = {},
  number = {},
  pages = {1-6},
  doi = {10.1109/SIBGRAPI62404.2024.10716316},
  issn = {1530-1834},
}

@ARTICLE{ill-2002,
  author={Baker, S. and Kanade, T.},
  journal={IEEE Transactions on Pattern Analysis and Machine Intelligence}, 
  title={Limits on super-resolution and how to break them}, 
  year={2002},
  volume={24},
  number={9},
  pages={1167-1183},
  doi={10.1109/TPAMI.2002.1033210}}

@article{song2020improved,
  title={Improved techniques for training score-based generative models},
  author={Song, Yang and Ermon, Stefano},
  journal={Advances in Neural Information Processing Systems (NeurIPS)},
  volume={33},
  pages={12438--12448},
  year={2020},
  booktitle = {Advances in Neural Information Processing Systems (NeurIPS)},
  doi={10.5555/3495724.3496767}
}

@InProceedings{chag-surv-2016,
author="Zhu, Shizhan
and Liu, Sifei
and Loy, Chen Change
and Tang, Xiaoou",
title="Deep Cascaded Bi-Network for Face Hallucination",
booktitle="European Conference on Computer Vision (ECCV)",
year="2016",
pages="614--630",
isbn="978-3-319-46454-1",
doi={\url{10.1007/978-3-319-46454-1_37}}
}

@article{li2022srdiff,
  title = {{SRDiff}: Single image super-resolution with diffusion probabilistic models},
  author = {Haoying Li and Yifan Yang and Meng Chang and Shiqi Chen and Huajun Feng and Zhihai Xu and Qi Li and Yueting Chen},
  year = {2022},
  journal = {Neurocomputing},
  volume = {479},
  pages = {47-59},
  doi = {10.1016/j.neucom.2022.01.029},
  issn = {0925-2312},
}

@inproceedings{song2021score,
  title={Score-based generative modeling through stochastic differential equations},
  author={Yang Song and Jascha Sohl-Dickstein and Diederik P. Kingma and Abhishek Kumar and Stefano Ermon and Ben Poole},
  booktitle = {International Conference on Learning Representations (ICLR)},
  month={May},
  year={2021},
  pages={1-36},
  doi={10.48550/arXiv.2011.13456}
}

@inproceedings{song2019generative,
  title = {Generative Modeling by Estimating Gradients of the Data Distribution},
  author = {Song, Yang and Ermon, Stefano},
  year = {2019},
  booktitle = {Advances in Neural Information Processing Systems (NeurIPS)},
  volume = {},
  pages = {1-13},
  doi={10.5555/3454287.3455354}
}

@inproceedings{sohl2015deep,
  title = {Deep Unsupervised Learning using Nonequilibrium Thermodynamics},
  author = {Sohl-Dickstein, Jascha and Weiss, Eric and Maheswaranathan, Niru and Ganguli, Surya},
  year = {2015},
  month = {},
  booktitle = {International Conference on Machine Learning (ICML)},
  volume = {},
  pages = {2256-2265},
  doi={10.48550/arXiv.1503.03585}
}

@inproceedings{ho2020denoising,
 author = {Ho, Jonathan and Jain, Ajay and Abbeel, Pieter},
 booktitle = {Advances in Neural Information Processing Systems (NeurIPS)},
 pages = {6840-6851},
 title = {Denoising Diffusion Probabilistic Models},
 volume = {33},
 year = {2020},
doi={10.5555/3495724.3496298}
}

@article{anderson1982reverse,
  title={Reverse-time diffusion equation models},
  author={Anderson, Brian DO},
  journal={Stochastic Processes and their Applications},
  volume={12},
  number={3},
  pages={313--326},
  year={1982},
  publisher={Elsevier},
  doi={10.1016/0304-4149(82)90051-5},
}

@INPROCEEDINGS{wang2021gfpgan,
  author={Wang, Xintao and Li, Yu and Zhang, Honglun and Shan, Ying},
  booktitle={IEEE/CVF Conference on Computer Vision and Pattern Recognition}, 
  title={Towards Real-World Blind Face Restoration with Generative Facial Prior}, 
  year={2021},
  volume={},
  number={},
  pages={9164-9174},
  doi={10.1109/CVPR46437.2021.00905}
}

@article{jolicoeur2021gotta,
  title={Gotta Go Fast When Generating Data with Score-Based Models},
  author={Jolicoeur-Martineau, Alexia and Li, Ke and Pich{\'e}-Taillefer, R{\'e}mi and Kachman, Tal and Mitliagkas, Ioannis},
  journal={arXiv preprint},
  year={2021},
  doi={10.48550/arXiv.2105.14080}
}

@inproceedings{vahdat2021score,
    title={Score-based Generative Modeling in Latent Space},
    author={Vahdat, Arash and Kreis, Karsten and Kautz, Jan},
    booktitle = {Advances in Neural Information Processing Systems (NeurIPS)},
    year={2021},
    pages = {11287-11302},
    volume = {34},
    doi={10.5555/3540261.3541124}
}

@inproceedings{cai2020learning,
  title={Learning gradient fields for shape generation},
  author={Cai, Ruojin and Yang, Guandao and Averbuch-Elor, Hadar and Hao, Zekun and Belongie, Serge and Snavely, Noah and Hariharan, Bharath},
  booktitle={European Conference on Computer Vision (ECCV)},
  pages={364-381},
  year={2020},
  doi={10.1007/978-3-030-58580-8\_22}
}

@inproceedings{song2022solving,
  title={Solving Inverse Problems in Medical Imaging with Score-Based Generative Models},
  author={Song, Yang and Shen, Liyue and Xing, Lei and Ermon, Stefano},
  year = {2022},
  booktitle = {International Conference on Learning Representations (ICLR)},
  pages = {1-18},
  doi={10.48550/arXiv.2111.08005},
}

@inproceedings{niu2020permutation,
  title = {Permutation Invariant Graph Generation via Score-Based Generative Modeling},
  author = {Chenhao Niu and Yang Song and Jiaming Song and Shengjia Zhao and Aditya Grover and Stefano Ermon},
  year = {2020},
  month = {Aug},
  booktitle = {International Conference on Artificial Intelligence and Statistics (AISTATS)},
  volume = {108},
  pages = {4474-4484},
  doi={10.48550/arXiv.2003.00638}
}

@article{vincent2011connection,
  title={A connection between score matching and denoising autoencoders},
  author={Vincent, Pascal},
  journal={Neural computation},
  volume={23},
  number={7},
  pages={1661--1674},
  year={2011},
  publisher={MIT Press},
doi={10.1162/NECO\_a\_00142}
}

@book{sarkka2019applied,
  title={Applied stochastic differential equations},
  author={S{\"a}rkk{\"a}, Simo and Solin, Arno},
  volume={10},
  year={2019},
  publisher={Cambridge University Press},
doi={10.1017/9781108186735}
}

@inproceedings{saharia2022photorealistic,
	title = {Photorealistic Text-to-Image Diffusion Models with Deep Language Understanding},
	author = {Saharia, Chitwan and Chan, William and Saxena, Saurabh and Lit, Lala and Whang, Jay and Denton, Emily and Ghasemipour, Seyed Kamyar Seyed and Ayan, Burcu Karagol and Mahdavi, S. Sara and Gontijo-Lopes, Raphael and Salimans, Tim and Ho, Jonathan and Fleet, David J and Norouzi, Mohammad},
	year = {2022},
	booktitle = {Advances in Neural Information Processing Systems (NeurIPS)},
	volume = {35},
	pages = {36479-36494},
    doi={10.5555/3600270.3602913}
}

@INPROCEEDINGS{zhang2018unreasonable,
  author={Zhang, Richard and Isola, Phillip and Efros, Alexei A. and Shechtman, Eli and Wang, Oliver},
  title={The Unreasonable Effectiveness of Deep Features as a Perceptual Metric}, 
  booktitle={IEEE/CVF Conference on Computer Vision and Pattern Recognition (CVPR)}, 
  year={2018},
  volume={},
  number={},
  pages={586-595},
  doi={10.1109/CVPR.2018.00068}
}

@INPROCEEDINGS{karras2019style,
  author={Karras, Tero and Laine, Samuli and Aila, Timo},
  booktitle={IEEE/CVF Conference on Computer Vision and Pattern Recognition (CVPR)}, 
  title={A Style-Based Generator Architecture for Generative Adversarial Networks}, 
  year={2019},
  volume={},
  number={},
  pages={4396-4405},
  doi={10.1109/CVPR.2019.00453}}

@book{sdeplaten,
author = {Kloeden, Peter and Platen, Eckhard},
year = {2011},
month = {Jan},
pages = {},
title = {Numerical Solution of Stochastic Differential Equations},
volume = {23},
isbn = {978-3-642-08107-1},
publisher={Springer},
doi = {10.1007/978-3-662-12616-5}
}

@article{neves2018quis,
  title={{QUIS-CAMPI}: an annotated multi-biometrics data feed from surveillance scenarios},
  author={Neves, Joao and Moreno, Juan and Proen{\c{c}}a, Hugo},
  journal={IET Biometrics},
  volume={7},
  number={4},
  pages={371--379},
  year={2018},
  publisher={Wiley Online Library},
doi={10.1049/iet-bmt.2016.0178}
}

@conference{visapp24sr,
author={Marcelo {dos Santos} and Joao Carlos Raposo Neves and Hugo Proen\c{c}a and David Menotti},
title={Defying Limits: Super-Resolution Refinement with Diffusion Guidance},
booktitle={International Conference on Computer Vision Theory and Applications (VISAPP)},
year={2024},
pages={426-434},
doi={10.5220/0012398900003660},
isbn={978-989-758-679-8},
issn={2184-4321},
}

@inproceedings{gao2023implicit,
  title={Implicit diffusion models for continuous super-resolution},
  author={Gao, Sicheng and Liu, Xuhui and Zeng, Bohan and Xu, Sheng and Li, Yanjing and Luo, Xiaoyan and Liu, Jianzhuang and Zhen, Xiantong and Zhang, Baochang},
  booktitle={IEEE/CVF Conference on Computer Vision and Pattern Recognition (CVPR)},
  pages={10021--10030},
  year={2023},
doi={10.1109/CVPR52729.2023.00966}
}

@inproceedings{dos2022face,
  title = {Face Super-Resolution Using Stochastic Differential Equations},
  author = {M. {dos Santos} and R. {Laroca} and R. O. {Ribeiro} and J. {Neves} and H. {Proen\c{c}a} and D. {Menotti}},
  year = {2022},
  month = {Oct},
  booktitle = {Conference on Graphics, Patterns and Images (SIBGRAPI)},
  volume = {},
  number = {},
  pages = {216-221},
  doi = {10.1109/SIBGRAPI55357.2022.9991799},
  issn = {1530-1834},
}

@article{yi2014learning,
  title={Learning face representation from scratch},
  author={Yi, Dong and Lei, Zhen and Liao, Shengcai and Li, Stan Z},
  journal={arXiv preprint},
  year={2014},
  doi={10.48550/arXiv.1411.7923}
}

@inproceedings{he2016deep,
  title={Deep residual learning for image recognition},
  author={He, Kaiming and Zhang, Xiangyu and Ren, Shaoqing and Sun, Jian},
  booktitle={IEEE/CVF Conference on Computer Vision and Pattern Recognition (CVPR)},
  pages={770--778},
  year={2016},
doi={10.1109/CVPR.2016.90}
}

@inproceedings{kim2022adaface,
  title={{AdaFace}: Quality Adaptive Margin for Face Recognition},
  author={Kim, Minchul and Jain, Anil K and Liu, Xiaoming},
  booktitle={IEEE/CVF Conference on Computer Vision and Pattern Recognition (CVPR)},
  year={2022},
  doi={10.1109/CVPR52688.2022.01201}
}

@inproceedings{meng2023distillation,
  title={On Distillation of Guided Diffusion Models},
  author={Meng, Chenlin and Rombach, Robin and Gao, Ruiqi and Kingma, Diederik and Ermon, Stefano and Ho, Jonathan and Salimans, Tim},
  booktitle={IEEE/CVF Conference on Computer Vision and Pattern Recognition (CVPR)},
  pages={14297-14306},
  year={2023},
  doi={10.1109/CVPR52729.2023.01374
}
}

@inproceedings{suin2024diffuse,
  title={Diffuse and restore: A region-adaptive diffusion model for identity-preserving blind face restoration},
  author={Suin, Maitreya and Nair, Nithin Gopalakrishnan and Pong Lau, Chun and Patel, Vishal M. and Chellappa, Rama},
  booktitle={IEEE/CVF Winter Conference on Applications of Computer Vision},
  pages={6343--6352},
  year={2024},
  doi={10.1109/WACV57701.2024.00622}
}

@inproceedings{zhang2023sine,
  title={{Sine}: Single image editing with text-to-image diffusion models},
  author={Zhang, Zhixing and Han, Ligong and Ghosh, Arnab and Metaxas, Dimitris and Ren, Jian},
  booktitle={IEEE/CVF Conference on Computer Vision and Pattern Recognition (CVPR)},
  pages={6027-6037},
  year={2023},
  doi={10.1109/CVPR52729.2023.00584}
}

@inproceedings{richter2023audio,
  title={Audio-visual speech enhancement with score-based generative models},
  author={Richter, Julius and Frintrop, Simone and Gerkmann, Timo},
  booktitle={ITG Conference on Speech Communication},
  pages={275-279},
  year={2023},
  doi={10.48550/arXiv.2306.01432}
}

@inproceedings{nascimento2022combining,
  title = {Combining Attention Module and Pixel Shuffle for License Plate Super-resolution},
  author = {V. {Nascimento} and R. {Laroca} and J. A. {Lambert} and W. R. {Schwartz} and D. {Menotti}},
  year = {2022},
  month = {Oct},
  booktitle = {Conference on Graphics, Patterns and Images (SIBGRAPI)},
  volume = {},
  number = {},
  pages = {228-233},
  doi = {10.1109/SIBGRAPI55357.2022.9991753},
  issn = {1530-1834},
}

@article{nascimento2024enhancing,
  title = {Enhancing License Plate Super-Resolution: A Layout-Aware and Character-Driven Approach},
  author = {V. {Nascimento} and R. {Laroca} and R. O. {Ribeiro} and W. R. {Schwartz} and D. {Menotti}},
  year = {2024},
  journal = {Conference on Graphics, Patterns and Images (SIBGRAPI)},
  volume = {},
  number = {},
  pages = {1-6},
  doi = {10.1109/SIBGRAPI62404.2024.10716303},
  issn = {1530-1834},
}

@article{tancik2020fourier,
  title={Fourier features let networks learn high frequency functions in low dimensional domains},
  author={Tancik, Matthew and Srinivasan, Pratul and Mildenhall, Ben and Fridovich-Keil, Sara and Raghavan, Nithin and Singhal, Utkarsh and Ramamoorthi, Ravi and Barron, Jonathan and Ng, Ren},
  journal={Advances in Neural Information Processing Systems (NeurIPS)},
  volume={33},
  pages={7537--7547},
  year={2020},
  doi={10.5555/3495724.3496356}
}

@InProceedings{ronneberger2015u,
author="Ronneberger, Olaf
and Fischer, Philipp
and Brox, Thomas",
editor="Navab, Nassir
and Hornegger, Joachim
and Wells, William M.
and Frangi, Alejandro F.",
title="U-Net: Convolutional Networks for Biomedical Image Segmentation",
booktitle="Medical Image Computing and Computer-Assisted Intervention -- MICCAI 2015",
year="2015",
pages="234--241",
isbn="978-3-319-24574-4",
doi={\url{10.1007/978-3-319-24574-4_28}}
}

@article{vaswani2017attention,
  title={Attention is all you need},
  author={Vaswani, Ashish and Shazeer, Noam and Parmar, Niki and Uszkoreit, Jakob and Jones, Llion and Gomez, Aidan N and Kaiser, {\L}ukasz and Polosukhin, Illia},
  journal={Advances in Neural Information Processing Systems (NeurIPS)},
  volume={30},
  year={2017},
  doi={10.5555/3295222.3295349}
}

@inproceedings{wang2017residual,
  title={Residual attention network for image classification},
  author={Wang, Fei and Jiang, Mengqing and Qian, Chen and Yang, Shuo and Li, Cheng and Zhang, Honggang and Wang, Xiaogang and Tang, Xiaoou},
  booktitle={IEEE/CVF Conference on Computer Vision and Pattern Recognition (CVPR)},
  pages={3156--3164},
  year={2017},
  doi={10.1109/CVPR.2017.683}
}

@INPROCEEDINGS{hermans2017defense,
  author={Yuan, Ye and Chen, Wuyang and Yang, Yang and Wang, Zhangyang},
  booktitle={IEEE/CVF Conference on Computer Vision and Pattern Recognition Workshops (CVPRW)}, 
  title={In Defense of the Triplet Loss Again: Learning Robust Person Re-Identification with Fast Approximated Triplet Loss and Label Distillation}, 
  year={2020},
  volume={},
  number={},
  pages={1454-1463},
  doi={10.1109/CVPRW50498.2020.00185}
}

@article{labach2019survey,
  title={Survey of dropout methods for deep neural networks},
  author={Labach, Alex and Salehinejad, Hojjat and Valaee, Shahrokh},
  journal={arXiv preprint},
  year={2019},
  doi={10.48550/arXiv.1904.13310}
}

\end{document}